\pdfoutput=1

\documentclass[11pt]{article}

\usepackage{acl}
\usepackage{graphicx}
\usepackage[export]{adjustbox}
\usepackage{caption}
\usepackage{subcaption}
\usepackage{amsmath}
\usepackage{times}
\usepackage{latexsym}
\usepackage{hyperref}       
\usepackage{url}
\usepackage{xcolor, soul}
\usepackage{graphicx}
\usepackage{booktabs}
\usepackage{multirow}
\usepackage{arabtex}
\usepackage{utf8}
\usepackage{pdflscape}
\setcode{utf8}
\usepackage{array}
\newcolumntype{H}{>{\setbox0=\hbox\bgroup}c<{\egroup}@{}}
\hypersetup{
colorlinks = true,
linkcolor = blue,
}
\usepackage[T1]{fontenc}

\usepackage[utf8]{inputenc}

\usepackage{microtype}

\title{ChatGPT for Arabic Grammatical Error Correction}

\author{\normalsize Sang Yun Kwon$^{1}$ ~ Gagan Bhatia $^{1}$ ~ El Moatez Billah Nagoudi$^{1}$ 
~\\ {\bf Muhammad Abdul-Mageed}$^{1,2}$\\
\normalsize $^{1}$Deep Learning \& Natural Language Processing Group,
  The University of British Columbia\\\normalsize  $^{2}$Department of Natural Language Processing \& Department of Machine Learning, MBZUAI\\ %
  \texttt{\normalsize \{skwon01@student.,gagan30@student.,moatez.nagoudi@,muhammad.mageed@\}ubc.ca}}

\begin{document}
\setcode{utf8}
\maketitle

\section*{~~~~~~~~~~~~~~~~~~~~~~~~~~~Abstract}
Recently, large language models (LLMs) fine-tuned to follow human instruction have exhibited significant capabilities in various English NLP tasks. However, their performance in grammatical error correction (GEC) tasks, particularly in non-English languages, remains significantly unexplored. In this paper, we delve into abilities of instruction fine-tuned LLMs in Arabic GEC, a task made complex due to Arabic's rich morphology. Our findings suggest that various prompting methods, coupled with (in-context) few-shot learning, demonstrate considerable effectiveness, with GPT-4 achieving up to $65.49$ F\textsubscript{1} score under expert prompting (approximately $5$ points higher than our established baseline). This highlights the potential of LLMs in low-resource settings, offering a viable approach for generating useful synthetic data for model training. Despite these positive results, we find that instruction fine-tuned models, regardless of their size, significantly underperform compared to fully fine-tuned models of significantly smaller sizes. This disparity highlights a substantial room for improvements for LLMs. Inspired by methods from low-resource machine translation, we also develop a method exploiting synthetic data that significantly outperforms previous models on two standard Arabic benchmarks. Our work sets new SoTA for Arabic GEC, with $72.19\%$ and $73.26$ F$\textsubscript{1}$ on the 2014 and 2015 QALB datasets, respectively.

\section{Introduction}\label{intro}
\begin{figure}[t]
  \centering
  \includegraphics[width=\linewidth]{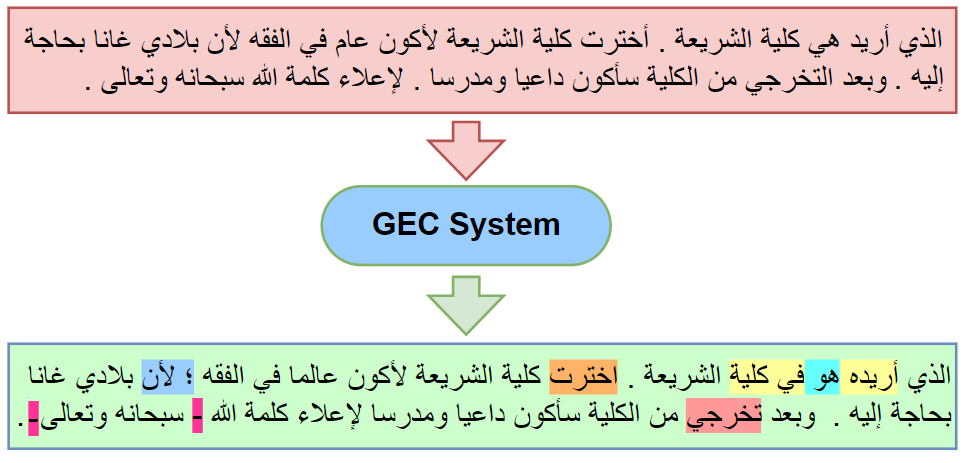}
\caption{An example of an Arabic GEC system showcasing six types of errors: \colorbox{cyan!30}{\textit{character replacement}}, \colorbox{yellow!50}{\textit{missing word}}, \colorbox{orange!35}{\textit{hamza error}}, \colorbox{blue!25}{\textit{missing punctuation}}, \colorbox{pink!55}{\textit{additional character}}, and \colorbox{violet!40}{\textit{punctuation confusion}}.}
\label{fig:trj_overview} 
\end{figure}

As interest in second language learning continues to grow, ensuring the accuracy and effectiveness of written language becomes increasingly significant for pedagogical tools and language evaluation~\cite{rothe-etal-2021-simple, tarnavskyi-etal-2022-ensembling}. A key component in this respect is grammatical error correction (GEC), a sub-area of natural language generation (NLG), which analyzes written text to automatically detect and correct diverse grammatical errors. Figure \ref{fig:trj_overview} shows an instance of GEC from~\cite{mohit2014first}.

Despite the growing attention to GEC, it is predominantly studied within the English language. One significant challenge in extending GEC systems to other languages is the lack of high-quality parallel data and benchmark datasets. In this work, our focus is on Arabic. Currently, the only available parallel data and benchmark datasets for Arabic GEC(AGEC) is the Qatar Arabic Language Bank (QALB)~\cite{mohit2014first, rozovskaya2015second}, highlighting the complexity of the task. Furthermore, Arabic, a language of complex grammar and rich morphological features, presents significant challenges to GEC. This further motivates our focus on Arabic in this work.

Non-English settings aside, the field of GEC has witnessed significant advancements specifically with the emergence of sequence-to-sequence (seq-2-seq) models~\cite{Chollampatt_Ng_2018,gong-etal-2022-revisiting} and sequence-to-edit models (seq-2-edit)~\cite{awasthi-etal-2019-parallel,omelianchuk2020gector} achieving SoTA results in the CONLL-2014 shared tasks \cite{ng-etal-2014-conll}. 

Although these models have achieved prominent performance, their efficacy relies heavily on large amounts of labeled data. Again, this presents challenges in low-resource scenarios. Recently, scaled up language models, \textit{aka} large language models (LLMs) have demonstrated remarkable potential in various NLP tasks.  Their core strength lies in their capacity to generalize across a wide range of languages and tasks, and in-context learning (ICL), enabling them to take on various NLP tasks once fed with only few examples (i.e., few–shot learning). A key component of this learning process is instruction fine-tuning, where these models are refined on a collection of tasks formulated as instructions~\cite{chung2022scaling}. This process amplifies the models' ability to respond accurately to such directives, reducing the need for few-shot examples~\cite{ouyang2022training, wei2022chain, sanh2021multitask}. With their unique features adeptly addressing the challenges of low-resource NLP scenarios, LLMs have emerged as promising candidates for NLP tasks in these scenarios. In our current study, we delve into the capabilities of LLMs, taking ChatGPT as our focus. We examine the effectiveness of various prompting strategies such as few-shot chain of thought (CoT) prompting~\cite{kojima2022large} and expert prompting~\cite{xu2023expertprompting}. Our research extends the realm of GEC research by concentrating on the unique challenges posed by Arabic, a complex and morphologically rich, low-resource language. Drawing upon the work of \citet{junczys-dowmunt-etal-2018-approaching}, we frame these challenges within the context of a low-resource MT task. We then carefully conduct a thorough comparison of the different methodologies employed in addressing GEC in Arabic. Our key contributions in this paper include:

\begin{enumerate}

\item[\textbf{1.}] We conduct a comprehensive investigation of the potential of LLMs, particularly focusing on ChatGPT, for tasks involving GEC in Arabic.

\item[\textbf{2.}] We provide a detailed examination of different prompting methods such as few-shot CoT and expert prompting, and an exploration into generating synthetic data with ChatGPT to complement the performance of transformer-based language models.

\item[\textbf{3.}] We further carry out in-depth and meaningful comparisons between several approaches (seq2seq, seq2-dit and instruction fine-tuning LLMs) in AGEC using the QALB 2014 and 2015 L1 benchmark dataset, allowing us to offer novel insights as to the utility of these  approaches on Arabic.


\end{enumerate}







The rest of this paper is organized as follows:
In Section~\ref{RW}, we review the related work on GEC, with a particular emphasis on Arabic. In Section~\ref{DE}, we describe available benchmark datasets for Arabic GEC and related evaluation metrics. In Section~\ref{BSES}, we describe our experimental setup; Section~\ref{LLM} outlines our experiments on LLMs. In Section~\ref{DA}, we introduce our seq-2-seq approach and Section~\ref{ST} discusses the sequence-2-edit approach. In \ref{AET}, we conduct a comprehensive analysis of error types using the ARETA~\cite{belkebir2021automatic}. Finally, in Section \ref{D}, we discuss our results, and in \ref{C}, we conclude the paper summarizing our contributions and outlining future research directions in the field of Arabic GEC.


\section{Related Work}\label{RW}

\paragraph{Progress in GEC.} 
Pre-trained Transformer-based models allowed for reframing GEC as a MT task~\cite{ng-etal-2014-conll, felice-etal-2014-grammatical, junczys2018approaching,grundkiewicz-etal-2019-neural}, leading to SoTA results. Meanwhile, sequence2edit methods cast the task as text-editing of input into an output~\cite{malmi2019encode,awasthi-etal-2019-parallel, omelianchuk2020gector}. These methods have simplified the complexity of model training while enhancing accuracy, especially in data-scarce scenarios. Furthermore, instruction fine-tuning~\cite{chung2022scaling} and various prompting techniques, such as the CoT~\cite{kojima2022large}, help optimize the performance of LLMs in the context of GEC. Finally, there is recent work that treats application of LLMs such as ChatGPT in GEC, demonstrating the effectiveness of these models. Further details regarding each of these approaches can be found in Appendix~\ref{app:RW_approaches}.

\paragraph{Arabic GEC.}
For Arabic GEC (AGEC), challenges stem from the complexity and morphological richness of Arabic. Arabic consists of a collection of diverse languages and 
dialectal varieties. Modern Standard Arabic (MSA) is a current standard variety of Arabic that is used in government and pan-arab media as well as education, alongside numerous regional dialects defined at the country or regional level \cite{abdul-mageed-etal-2020-nadi}. The inherent ambiguity of Arabic at the orthographic, morphological, syntactic, and semantic levels makes AGEC particularly challenging. Optional use of diacritics further introduces orthographic ambiguity~\cite{belkebir2021automatic}, making AGEC even harder.

Despite these hurdles, progress has been made in AGEC. For example, the QALB-2014 and 2015 shared task \cite{mohit2014first, rozovskaya-etal-2015-second}, released annotated datasets of comments and documents written by native (L1) and Arabic learner (L2) speakers. More  recently,the ZAEBUC corpus~\cite{habash-palfreyman-2022-zaebuc} a GEC corpus of essays written by first year university students in Zayed University, UAE. In terms of model development, innovative approaches have been introduced.~\citet{watson2018utilizing} develop the first character-level seq2seq model that achieved SoTA results on AGEC L1 data. \citet{solyman2022automatic,SOLYMAN2021303} design a model that utilizes a dynamic linear combination and EM routing algorithm with a seq2seq Transformer. Convolutional neural network (CNN) have also been applied to AGEC, using unsupervised noise injection techniques to generate synthetic parallel data~\cite{solyman2022automatic,SOLYMAN2021303, SOLYMAN2023101572}. In spite of this progress, no work has considered the utility of employing ChatGPT (or any LLM in general) for AGEC. Nor has been significant work on exploring synthetic data generation, including from LLMs or adopting more diverse machine learning approaches, been carried out. Our research fills this existing gap.


\section{Datasets \& Evaluation}\label{DE}

\subsection{Datasets}
In this study, we make use of the 2014~\cite{mohit2014first} and 2015~\cite{rozovskaya-etal-2015-second} QALB Shared Task datasets to evaluate the performance of our various models. Both datasets make use of the QALB corpus, a manually corrected collection of Arabic texts. These texts include online commentaries from Aljazeera articles in MSA by L1 speakers, as well as texts produced by L2 learners of Arabic. Both the QALB 2014 and 2015 dataset are split into training (Train), development (Dev), and test (Test) sets based on their annotated dates. QALB 2014 consists of  $19,411$ sentences, $1,017$ sentences, and $968$ sentences for the respective Train, Dev, and Test splits. QALB 2015 is an extension of the first 2014 shared task, including L1 commentaries and L2 texts that cover different genres and error types. For the purposes of our study, we exclusively utilize the L1 test set (2015), as we focus on sentence-level AGEC, where L2 test sets are document-level. The parts of the 2015 dataset we use are comprised of $300$ sentences for Train, $154$ sentences for Dev, and $920$ sentences for the L1 Test set. 
Statistics of these datasets are in Table~\ref{tab:DatasetDescriptions}.

\begin{table}[]
\centering
\resizebox{\columnwidth}{!}{
\begin{tabular}{lllllll}
\toprule
\textbf{Dataset} & \textbf{Split} & \textbf{Lines} & \textbf{Words} & \textbf{Err. \%} & \textbf{Level} & \textbf{Domain}\\
\midrule
\multirow{3}{*}{\textbf{QALB-2014}}
 & Train & 19,411 & 1,021,165 & 30\% & Native & Comments\\
 & Dev & 1,017 & 53,737 & 31\% & Native & Comments\\
 & Test & 968 & 51,285 & 32\% & Native & Comments\\
\midrule
\multirow{4}{*}{\textbf{QALB-2015}}
 & Train & 310 & 43,353 & 30\% & L2 & Essays\\
 & Dev & 154 & 24,742 & 29\% & L2 & Essays\\
 & Test-L2 & 158 & 22,808 & 27\% & L2 & Essays\\
 & Test-L1 & 920 & 48,547 & 29\% & Native & Comments\\
\bottomrule
\end{tabular}}
\caption{Dataset descriptions and statistic for train, development (dev) and test data.}
\label{tab:DatasetDescriptions}
\end{table}




\subsection{Evaluation Metric}
For evaluation, we utilize the overlap-based metric MaxMatch (M$^2$)~\cite{dahlmeier-ng-2012-better}, which aligns source and hypothesis sentences based on Levenshtein distance , selecting maximal matching edits, scoring the precision (P), recall (R), and F\textsubscript{1} measure. Moreover, in alignment with recent works on GEC, we also report the F\textsubscript{0.5} score as another important evaluation metric, a variation of the F\textsubscript{1} score that place twice as much weight on precision than on recall. This reflects a general consensus that precision holds greater importance than comprehensive error correction in GEC systems.

\noindent{\textbf{Normalisation methods.}}  Following the QALB shared task's evaluation method, we report system performance across three different categories, targeting distinct types of mistakes. Namely, we assess the system on normalized text with (1) Alif/Ya errors removed, (2) text without punctuation, and (3) text devoid of both Alif/Ya errors and punctuation. Although we primarily focus on the \textit{'Exact Match'} results for analysis and discussion,  scores for most experiments are provided in our Appendixes. Examples of text under each setting, along with the comprehensive results, can be found in the Appendix~\ref{errors_normaliszation}.




\section{Baseline and Experimental Setup} \label{BSES}
Our baseline settings include AraBart~\cite{eddine2022arabart} and AraT5~\cite{nagoudi2021arat5}, text-to-text transformer-based models specifically tailored for Arabic tasks. We also evaluate the performance of the mT0~\cite{muennighoff2022crosslingual} and mT5~\cite{xue2020mt5} variants of the T5 model~\cite{raffel2020exploring}, both of which are configured for multilingual tasks. 

For our experiments, we fine-tune each models for 15 epochs. We employ a learning rate of 5e-5 and a batch size of 32, then picking the best-performing model on our Dev data before blind-testing on Test. 

\section{LLMs and Prompting Techniques}\label{LLM}
This section outlines our experiment designed to instruction fine-tune LLMs and explore different prompting methods for ChatGPT in the context of GEC. We begin by experimenting with various prompting strategies using ChatGPT, comparing its performance against smaller LLMs and our listed baselines. We evaluate the performance of ChatGPT-3.5 Turbo and ChatGPT-4 using the official API, under two distinct prompting approaches: \textbf{\textit{Few-shot CoT}}~\cite{fang2023chatgpt} and \textbf{\textit{Expert Prompting}}~\cite{xu2023expertprompting}. We now describe our prompting strategies.

\noindent\subsection{ChatGPT Prompting}

\paragraph{Preliminary experiment.} 
Initially, we experiment with a diverse set of prompt templates to assess ChatGPT’s capabilities in zero-shot learning as well as two aspects of few-shot learning: vanilla few-shot and few-shot CoT ~\cite{fang2023chatgpt}. We also experiment with prompts in both English and Arabic. However, we discover that the responses from these prompt templates contain extraneous explanations and are disorganized, necessitating substantial preprocessing for compatibility with the M$^2$ scorer. This problem was particularly notable in the zero-shot and Arabic prompt setups, which failed to yield output to automatically evaluate. 

\begin{figure*}[]
    \centering
    \includegraphics[width=\textwidth]{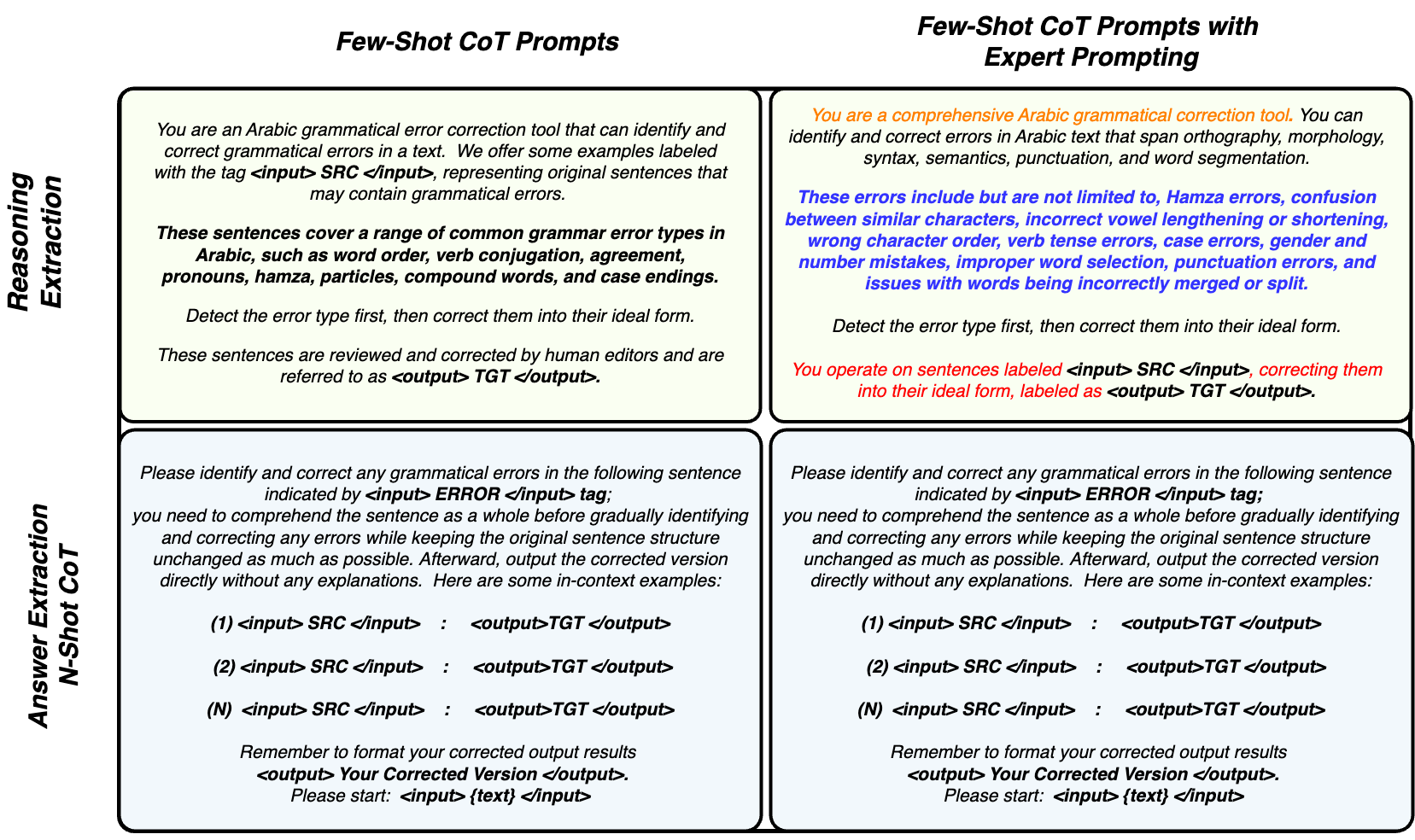}
    \caption{An illustration of Few-Shot CoT and Expert Prompts applied to CharGPT for Grammatical Error Correction.}
    \label{fig:x cubed graph}
\end{figure*}

\paragraph{Few-shot CoT.} 
Adopting the few-shot CoT prompt design strategy from~\citet{kojima2022large} and~\citet{fang2023chatgpt}, we implement a two-stage approach. Initially, we engage in \textit{'reasoning extraction'}, prompting the language model to formulate an elaborate reasoning pathway. This is followed by an \textit{'answer extraction'} phase, where the reasoning text is combined with an answer-specific trigger sentence to form a comprehensive prompt. These directives include tailored prompts that position the model as an Arabic GEC tool. In the few-shot CoT setting, we include labeled instances from the development set in our prompts to implement in-context learning (ICL), facilitating learning from examples~\cite{brown2020language}. This approach involves the use of erroneous sentences, indicated by \colorbox{red!10}{\textit{<input> SRC </input>}}, along with their corrected versions, coded by \colorbox{green!10}{\textit{<output> TGT </output>}}.

\paragraph{Expert prompting.}
\citet{xu2023expertprompting} introduces a novel strategy, which leverages the expert-like capabilities of LLMs. This method involves assigning expert personas to LLMs, providing specific instructions to enhance the relevance and quality of the generated responses. Following the framework proposed by \citet{xu2023expertprompting}, we ensure that our Arabic GEC correction tool exhibits three key characteristics: being \textcolor{orange}{\textit{\textbf{distinguished}}}, \textcolor{blue}{\textit{\textbf{informative}}}, and \textcolor{red}{\textit{\textbf{automatic}}} during the \textit{'reasoning extraction'} stage of our prompt. To achieve this, we curate a distinct and informative collection of various error types rooted in the dataset, referencing the taxonomy of the Arabic Learner Corpus~\cite{alfaifi2012arabic}. Then we prompt to automate the system by instructing it to operate on sentence labeled with \colorbox{red!10}{\textit{<input>}} and \colorbox{green!10}{\textit{<output>}} tags. Details of both prompts can be found in Figure \ref{fig:x cubed graph}.

\subsection{ChatGPT Results.}Table~\ref{tab:ChatGPT_eval} presents the performance of ChatGPT under different prompting strategies in comparison to the baseline settings. Noticeably, we observe improvements particularly as we progress from the one-shot to five-shot configurations under both the few-shot CoT and expert prompting (EP) strategies. Under the CoT prompt, the F\textsubscript{1.0} score for ChatGPT increased from $53.59$ in the one-shot setting to $62.04$ in the five-shot setting. A comparable upward trend was also evident for the EP strategy, with the F\textsubscript{1.0} score improving from $55.56$ in the one-shot setup to $63.98$ in the five-shot setup. Furthermore, among all the ChatGPT trials, the three-shot and five-shot configurations of ChatGPT-4, under the CoT strategy, yield the highest scores. These configurations achieve F\textsubscript{1.0} scores of $63.88$ and $65.49$, respectively.

\noindent\subsection{Instruction-Finetuning LLMs}
InstructGPT ~\cite{ouyang2022training} and studies on ChatGPT 2, highlight the potential of LLMs to excel in various downstream tasks just by leveraging a few examples as instructions. Further advancements have been realized by fine-tuning language models on a compilation of tasks presented as instructions, enhancing models' responsiveness, and minimizing the need for few-shot exemplars \cite{chung2022scaling}.

In this study, we extend the application of instruction-finetuning to AGEC tasks across a broad range of models that range in size, including LLaMA-7B \cite{touvron2023llama}, Vicuna-13B \cite{vicuna2023}, Bactrian-X$_{\textit{bloom}}$-7B~\cite{bactrian}, Bactrian-X$_{\textit{llama}}$-7B~\cite{bactrian}.

\paragraph{LLM finetuning.} 
To instruct fine-tune relatively large language models, \textit{henceforth} simply LLMs, we first pre-train LLMs on the translated Alpaca dataset~\footnote{We translate Alpaca  datasets using NLLB MT model~\cite{costa2022no}} to help our model gain deeper understanding of the Arabic language and its complexities. Following this, we further fine-tune the models on our GEC dataset, targeting specifically the task of GEC~\cite{alpaca}. Then, we employ well-structured task instructions and input prompts, enabling the models to take on GEC tasks. Each model is assigned a task, given an instruction and an input for output generation. A detailed illustration of the instructions utilized for the models can be found in Appendix \ref{app:IF}.

\begin{table}[t]
\centering
\resizebox{\columnwidth}{!}{%
\begin{tabular}{llcccc}\toprule
\multirow{2}{*}{\textbf{Settings}} &\multirow{2}{*}{\textbf{Models}} &\multicolumn{4}{c}{\textbf{Exact Match}} \\\cmidrule{3-6}
& &\textbf{P} &\textbf{R} &\textbf{F\textsubscript{1.0}} &\textbf{F\textsubscript{0.5}} \\ \toprule 
\multirow{5}{*}{\textbf{Baselines}} 
&mT0 & $69.16$ & $52.63$ & $59.78$ & $65.07$ \\    
&mT5 &$68.99$ &$52.22$ &$59.45$ &$64.83$ \\
&AraBART &$70.71$ &$60.46$ &$65.18$ &$68.39$ \\
&AraT5 &$\mathbf{73.04}$ &$\mathbf{63.09}$ &$\mathbf{67.70}$ &$\mathbf{70.81}$ \\
\midrule
\multirow{3}{*}{\textbf{+ CoT}} 
&ChatGPT (1-shot) &$58.71$ &$49.29$ &$53.59$ &$56.55$ \\
&ChatGPT (3-shot) &$64.60$ &$60.37$ &$62.41$ &$63.71$ \\
&ChatGPT (5-shot) &$64.70$ &$59.59$ &$62.04$ &$63.61$ \\
\midrule
\multirow{3}{*}{\textbf{+ EP}} 
&ChatGPT (1-shot) &$60.49$ &$51.37$ &$55.56$ &$58.42$ \\
&ChatGPT (3-shot) &$65.83$ &$61.41$ &$63.54$ &$64.90$ \\
&ChatGPT (5-shot) &$66.53$ &$61.62$ &$63.98$ &$65.49$ \\
\midrule
\multirow{3}{*}{\textbf{+ CoT}} 
&GPT4 (1-shot) \textsuperscript{$*$} & $-$  & $-$  & $-$  & $-$  \\
&GPT4 (3-shot) &$69.31$ &$59.24$ &$63.88$ &$67.03$ \\
&GPT4 (5-shot) &$69.46$ &$61.96$ &$65.49$ &$67.82$ \\
\toprule
\end{tabular}%
}
\caption{Performance of ChatGPT under different prompting strategies.\textsuperscript{$*$}Results for GPT4 1-shot are not included due to the high cost of producing these results, and a pattern has already been established showing that scores increase with the number of N-shot examples} \label{tab:ChatGPT_eval}
\end{table}

\begin{figure}[t]
\centering
\includegraphics[width=1.0\columnwidth]{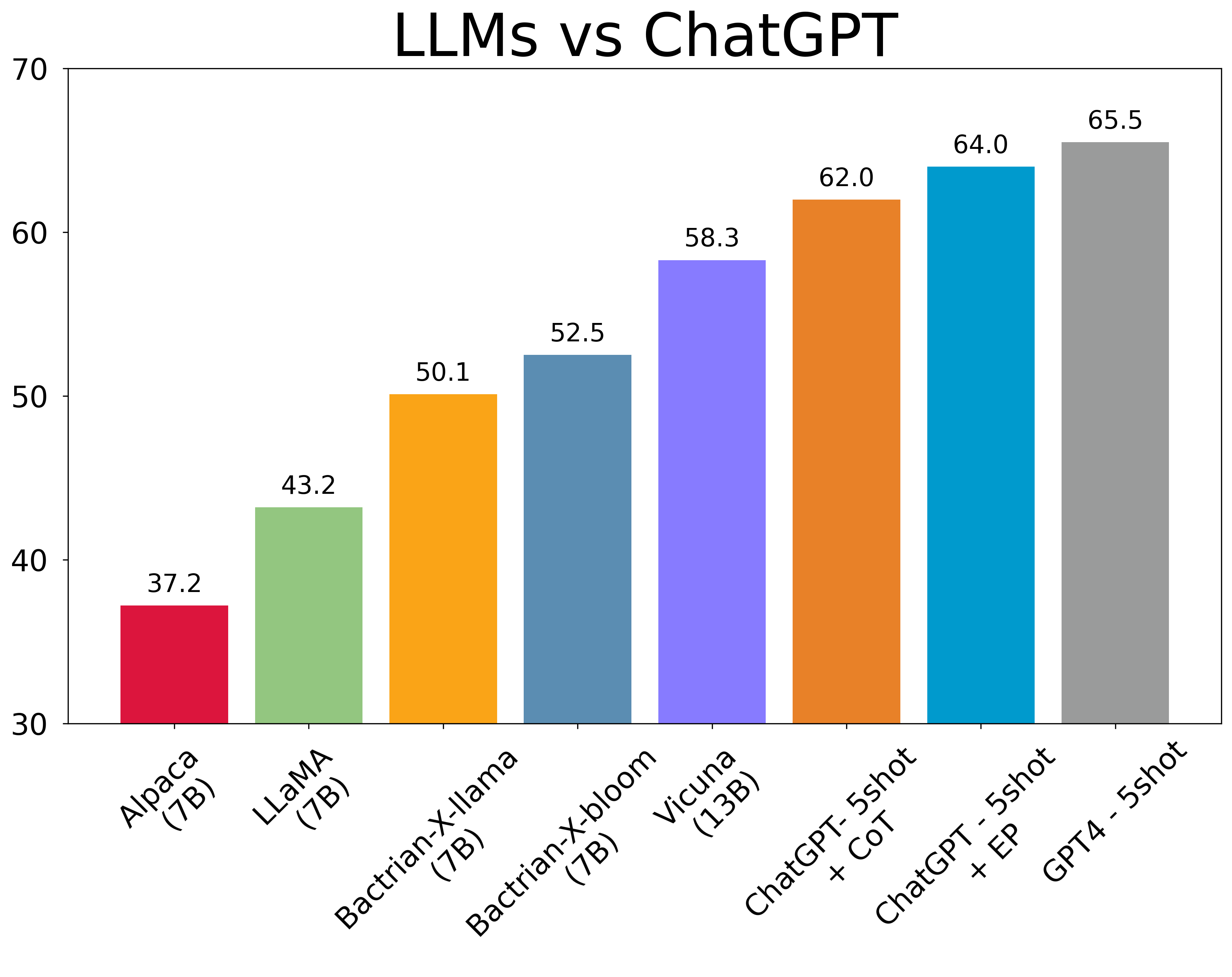}
\caption{Performance of LLMs compared to ChatGPT.}
\label{fig:gptllm}
\end{figure}

\paragraph{LLM Results.} 
As shown in Figure~\ref{fig:gptllm}, larger models such as Vicuna-13B and models trained on multilingual data like Bactrian-X$_{\textit{llama}}$-7B, and  Bactrian-X$_{\textit{bloom}}$-7B exhibit an overall trend of better performance, achieving F\textsubscript{1} scores of $58.30$, $50.1$, and $52.5$, respectively. Despite these improvements, it is noteworthy that all these LLMs fall short of ChatGPT's performance in the AGEC tasks, reaffirming ChatGPT's superior ability in this context.

\section{Data Augmentation} \label{DA}
Motivated by the significant improvements observed in low-resource GEC tasks in languages such as German, Russian, and Czech through synthetic data creation \cite{flachs-etal-2021-data}, and recognizing the recent efforts to develop synthetic data for AGEC \cite{SOLYMAN2021303}, we experiment with three distinctive data augmentation methods, evaluating the efficacy of each method in complementing performance of seq-2-seq models. 

\paragraph{ChatGPT as corruptor.}
With slight adaptation to our original  prompt, we engage ChatGPT as an AI model with the role of introducing grammatical errors into Arabic text to generate artificial data. We randomly sample $10,000$ correct sentences from the original training set and prompt ChatGPT to corrupt these, creating a parallel dataset. In order to ensure a varied range of error types, we adopt the taxonomy put forth by the Arabic Learner Corpus~\cite{alfaifi2012arabic}.

\paragraph{Token noising and error adaptation.}
Adopting techniques known as \textit{token noising}~\cite{xie-etal-2018-noising} and \textit{error adaptation}~\cite{junczys-dowmunt-etal-2018-approaching}, we generate artificial data by introducing random alterations and matching the error rates of the original benchmark dataset, in table ~\ref{tab:DataErrorTypePercentages}. For token noising, random character-level and word-level changes are introduced into clean texts, creating a parallel dataset. These changes include random character manipulations, word separations, space adjustments, Arabic text normalization, and inserting or removing punctuation. To ensure domain compatibility with the original benchmark dataset, we use commentaries from the same newspaper domain as our clean inputs and adjust sentence lengths to align with the benchmark dataset. 

\begin{table}[h!]
\centering
\resizebox{1\columnwidth}{!}{
\begin{tabular}{lccccccc}
\toprule
 & \textbf{Edit} & \textbf{Add} & \textbf{Merge} & \textbf{Split} & \textbf{Delete} & \textbf{Move} & \textbf{Other}\\
\midrule
\textbf{Train} & 55.34\% & 32.36\% & 5.95\% & 3.48\% & 2.21\% & 0.14\% & 0.50\%\\
\textbf{Dev} & 53.51\% & 34.24\% & 5.97\% & 3.67\% & 2.03\% & 0.08\% & 0.49\%\\
\textbf{Test} & 51.94\% & 34.73\% & 5.89\% & 3.48\% & 3.32\% & 0.15\% & 0.49\%\\
\bottomrule
\end{tabular}}
\caption{Error type statistics for training, development, and test Sets.}
\label{tab:DataErrorTypePercentages}
\end{table}

\paragraph{Reverse noising.}
We adopt a \textit{reverse noising} approach~\cite{xie-etal-2018-noising}, training a reverse model that converts clean sentences \textit{Y} into noisy counterparts \textit{X}. This involves implementing a standard beam search to create noisy targets $\hat{Y}$ from clean input sentences \textit{Y}. Our approach incorporates two types of reverse models: the first trains both QALB-2014 and 2015 datasets, and the second makes use of a parallel dataset generated by '\textit{ChatGPT as corruptor}'. Subsequently, we produce two parallel datasets by inputting clean 'in-domain' and 'out-of-domain' examples from each reverse model. In this context, the 'in-domain' dataset refers to news article commentaries, the same as the original training data, and 'out-of-domain' refers to any Arabic sentences.

\begin{figure*}[]
\centering
\includegraphics[width=\textwidth]{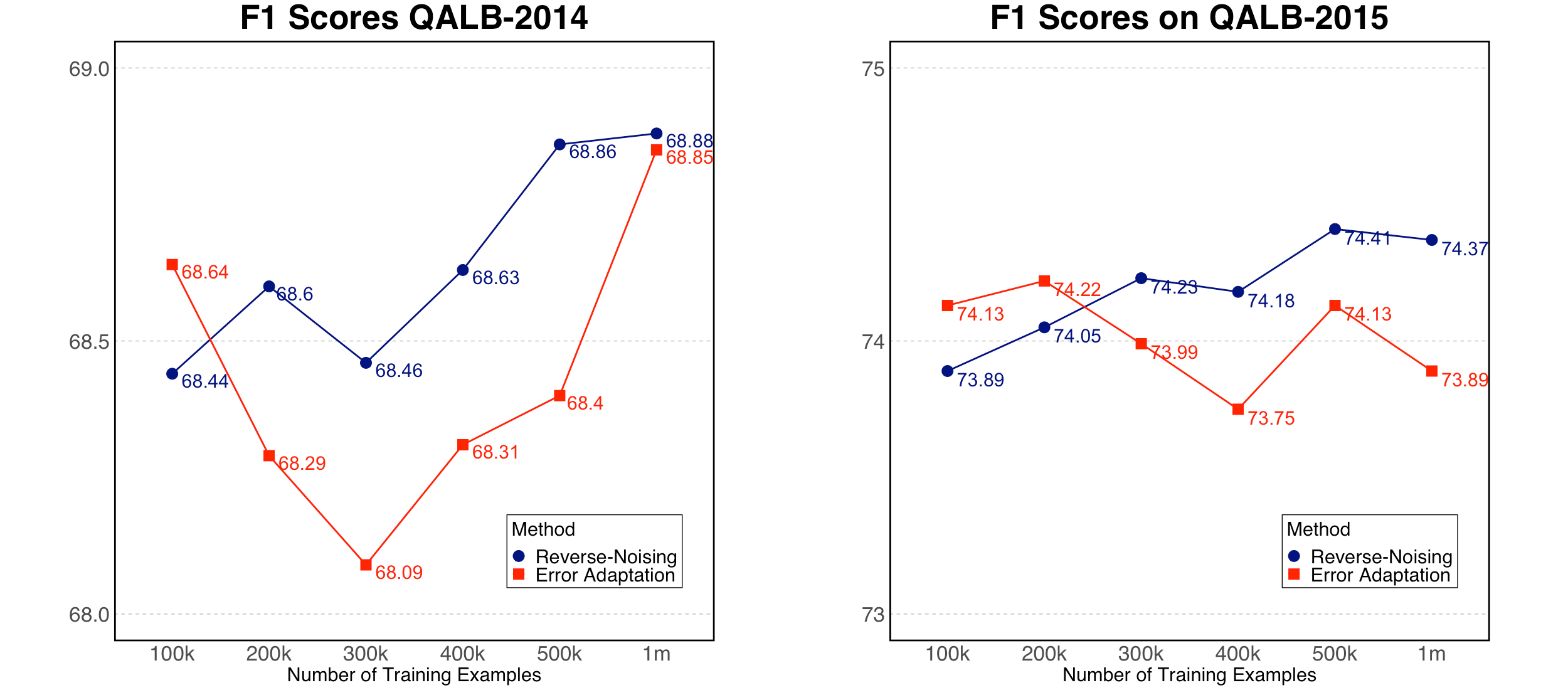}
\caption{Scores trained on 100k to 1m sentences of training data}
\label{fig:DA}
\end{figure*}

\begin{figure*}[!h]
\centering
\includegraphics[width=\textwidth]{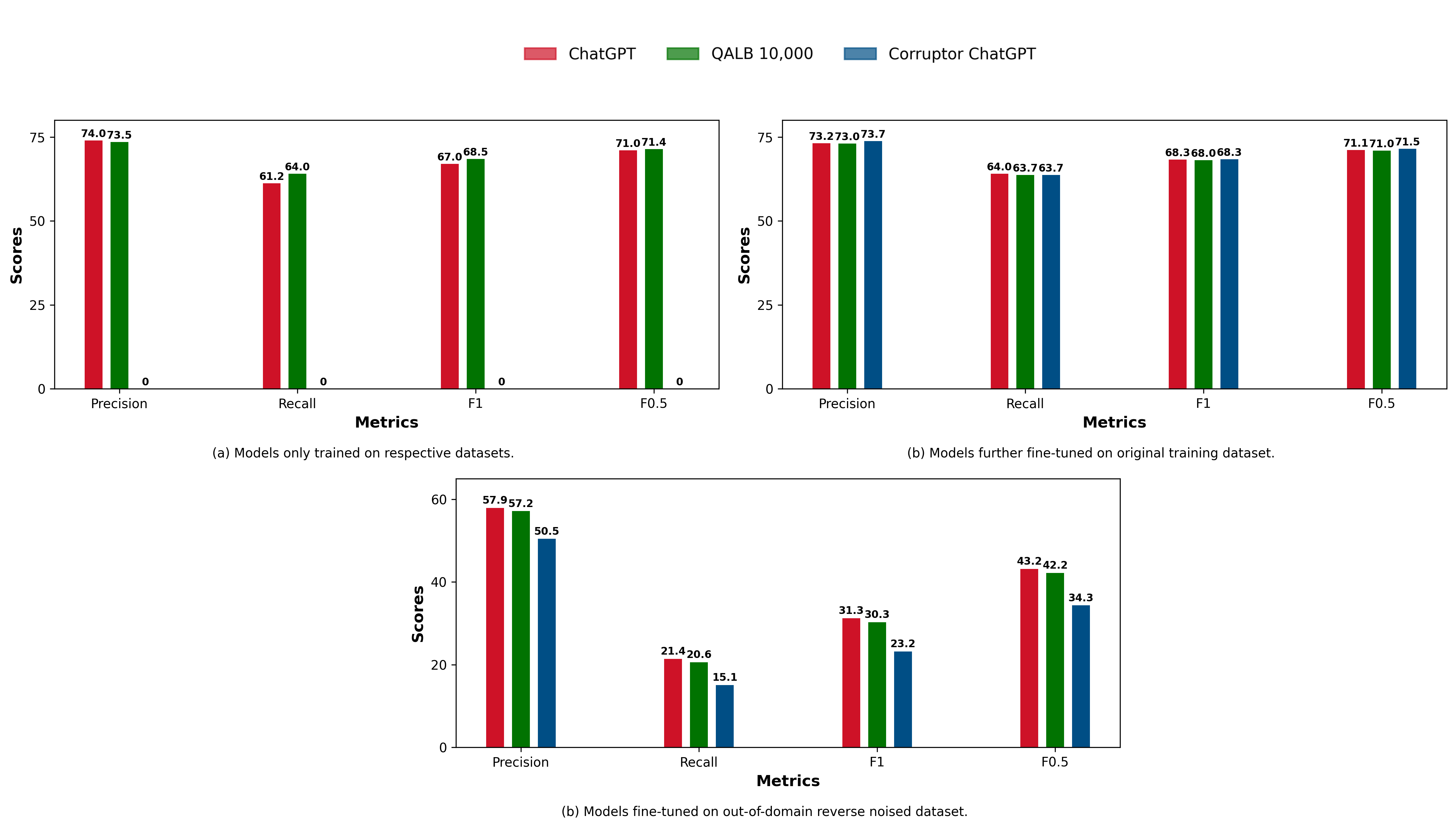}
\caption{Scores of models fine-tuned on $10,000$ parallel sentences from different sources: original training data, 'ChatGPT as Corruptor', and reverse noising on the ChatGPT dataset.}
\label{fig:three}
\end{figure*}

\paragraph{Data augmentation evaluation.}
To evaluate the efficacy of ChatGPT in generating artificial data, we select $10,000$ parallel sentences generated through `\textit{ChatGPT as corruptor}', $10,000$ examples from the parallel dataset from reverse noising on the ChatGPT dataset as well as $10,000$ parallel sentences from the original training set. We then further fine-tune each of the configurations on the original training dataset and  the `out-of-domain' reverse noised dataset, aiming to assess whether these artificially created datasets can replace the gold standard training set. Figure~\ref{fig:three} outlines the results. In our initial exploration, fine-tuning the AraT5 model exclusively on $10,000$ samples, ChatGPT achieves an F\textsubscript{1} of $67.00$, scoring slightly below the original QALB 2014 training data ($68.45$). Subsequently, when further fine-tuned on the original training set, our model (F\textsubscript{1} score at $68.39$) is on par with the AraT5 model further fine-tuned on the equivalent-sized gold dataset (F\textsubscript{1} score at $68.05$) . This confirms the utility of ChatGPT for generating synthetic data. Conversely, when we further fine-tune the model on $10,000$ out-of-domain examples, its performance drops significantly (F\textsubscript{1} =$44.65$). This underscores the importance of relevant and high-quality synthetic data over randomly generated samples.

We scale our data augmentation experiments comparing the 'token noising' and error adaptation' and 'reverse-noising'. Results, outlined in Figure~\ref{fig:DA}, show consistent improvement over the baseline. The \textit{'token noising and error adaptation'} method helps improve the F\textsubscript{1} scores, with a range of $68.09$ to $68.85$, attaining optimal performance with the one million dataset size. Similarly, the \textit{'reverse noising'} method, yielding scores from $68.44$ to $68.88$, also reaches its peak performance at the one million datasets. Both methods exhibit similar performance trends when tested on the QALB-2015 dataset.

\section{Sequence Tagging Approach} \label{ST}
In this section, we detail our methods to adapt the GECToR model~\cite{omelianchuk2020gector} for experimenting with the sequence-to-edit approach.

\paragraph{Token level transformations.}
We first apply token-level transformation to recover the target text by applying them to the source tokens. \textit{\textbf{`Basic-transformations'}} are applied to perform the most common token-level edit operations, such as keeping the current token unchanged \textit{($KEEP$)}, deleting current token \textit{($DELETE$)}, appending new token t\_\textsubscript{$1$} next to the current token x\textsubscript{i} \textit{($APPEND$\_~t\textsubscript{$1$})} or replacing the current token x\textsubscript{i} with another token t\_\textsubscript{$2$} \textit{($REPLACE$\_~t\textsubscript{$2$})}. To apply tokens with more task specific operations we employ \textit{\textbf{`g-transformations'}} such as  the \textit{($MERGE$)} tag to merge the current token and the next token into a single one. Edit space after applying token-level transformations results in \textit{KEEP} ($725$K op), \textit{DELETE} ($13$K op), \textit{MERGE} ($5.7$K op), \textit{APPEND\_t\textsubscript{$1$}} ($75$K op), and \textit{REPLACE\_t\textsubscript{2}} ($201$K op) tags.  As some corrections in a sentence depend on others, applying edit sequences once may not be enough to fully correct the sentence. To address this issue, GECToR employs an iterative correction approach from ~\citet{awasthi-etal-2019-parallel}. However, in our experiments, we find that the iterative correction approach does not result in any tangible improvement. Therefore, we set our iterations to $3$.


\paragraph{Preprocessing and fine-tuning.}
We start the preprocessing stage by aligning source tokens with target subsequences, preparing them for token-level transformations. Subsequently, we fine-tune ARBERT v2  ~\cite{elmadany2022orca} and MARBERT v2\cite{abdul-mageed-etal-2021-arbert} on the preprocessed data. Then we employ a three step training procedure: an initial pre-training phase using artificially generated sentences with errors; a fine-tuning phase that exclusively uses sentences containing errors; and a final refinement phase that employs a combination of sentences, both with and without errors. 

\paragraph{Sequence tagging evaluation.}
Outlined in Table~\ref{tab: Gector_Results}, ARBERT v2 and MARBERT v2, exhibit high precision , with ARBERT v2's three-step training scoring the highest precision at $74.39$. However, relatively lower recall scores indicate challenges in ability of the two models to detect errors. The implementation of a three-stage training approach yielded mixed results: while accuracy improves, recall scores decrease, leading to a drop in the overall F\textsubscript{1} score (by $0.36$ for ARBERT v2 and $1.10$ for MARBERT v2, respectively). Consequently, all models fall behind the 'seq2seq' models in performance. However, both ARBERT v2 and MARBERT v2 surpass 'mT0' and 'mT5' in terms of F\textsubscript{0.5} scores highlighting their abilities in correcting errors with precision.

\begin{table}[]
\centering
\renewcommand{\arraystretch}{1.2}
\resizebox{\columnwidth}{!}{%
\begin{tabular}{llcccc|cccc}\toprule
\multirow{2}{*}{\textbf{Settings}} &\multirow{2}{*}{\textbf{Models}} &\multicolumn{4}{c}{\textbf{QALB-2014}} &\multicolumn{4}{c}{\textbf{QALB-2015}}\\\cmidrule{3-6}\cmidrule{7-10}
& &\textbf{P} &\textbf{R} &\textbf{F\textsubscript{1.0}} &\textbf{F\textsubscript{0.5}} &\textbf{P} &\textbf{R} &\textbf{F\textsubscript{1.0}} &\textbf{F\textsubscript{0.5}}\\\toprule
\multirow{4}{*}{\textbf{Encoder-Decoder}} 
&mT0 &$69.16$ &$52.63$ &$59.78$ &$65.07$ &$67.61$ &$58.41$ &$62.67$ &$65.55$ \\
&mT5 &$68.99$ &$52.22$ &$59.45$ &$64.83$ &$67.43$ &$57.30$ &$61.95$ &$65.13$ \\
&AraBART &$70.71$ &$60.46$ &$65.18$ &$68.39$ &$67.95$ &$65.62$ &$66.76$ &$67.47$ \\
&AraT5 &$73.04$ &$\mathbf{63.09}$ &$\mathbf{67.70}$ &$\mathbf{70.81}$ &$72.41$ &$\mathbf{74.12}$ &$\mathbf{73.26}$ &$\mathbf{72.75}$ \\\midrule
\multirow{4}{*}{\textbf{Encoder-Only}} &ARBERTv2 &$73.89$ &$48.33$ &$58.43$ &$66.82$ &$73.10$ &$55.40$ &$63.03$ &$68.70$ \\
&ARBERTv2 (3-step) &$\mathbf{74.39}$ &$47.62$ &$58.07$ &$66.87$ &$\mathbf{74.20}$ &$53.80$ &$62.37$ &$68.96$ \\
&MARBERTv2 &$73.53$ &$48.21$ &$58.24$ &$66.54$ &$72.90$ &$54.90$ &$62.63$ &$68.41$ \\
&MARBERTv2 (3-step) &$74.21$ &$46.45$ &$57.14$ &$66.29$ &$74.00$ &$52.70$ &$61.56$ &$68.46$ \\\bottomrule
\end{tabular}%
}
\caption{F\textsubscript{1} and F\textsubscript{0.5} scores of sequence tagging approach for QALB-2014 and QALB-2015}\label{tab: Gector_Results}
\end{table}

\section{Error Analysis} \label{AET} 
\subsection{Error Types.} Using the Automatic Error Type Annotation (ARETA) tool~\cite{belkebir2021automatic} we examine the performance of error types of our developed models. In the following, we briefly describe the different error's types included in QALB, as well as the normalization methods used to evaluate the model performances.

\begin{figure*}[h]
\centering
\includegraphics[width=\textwidth]{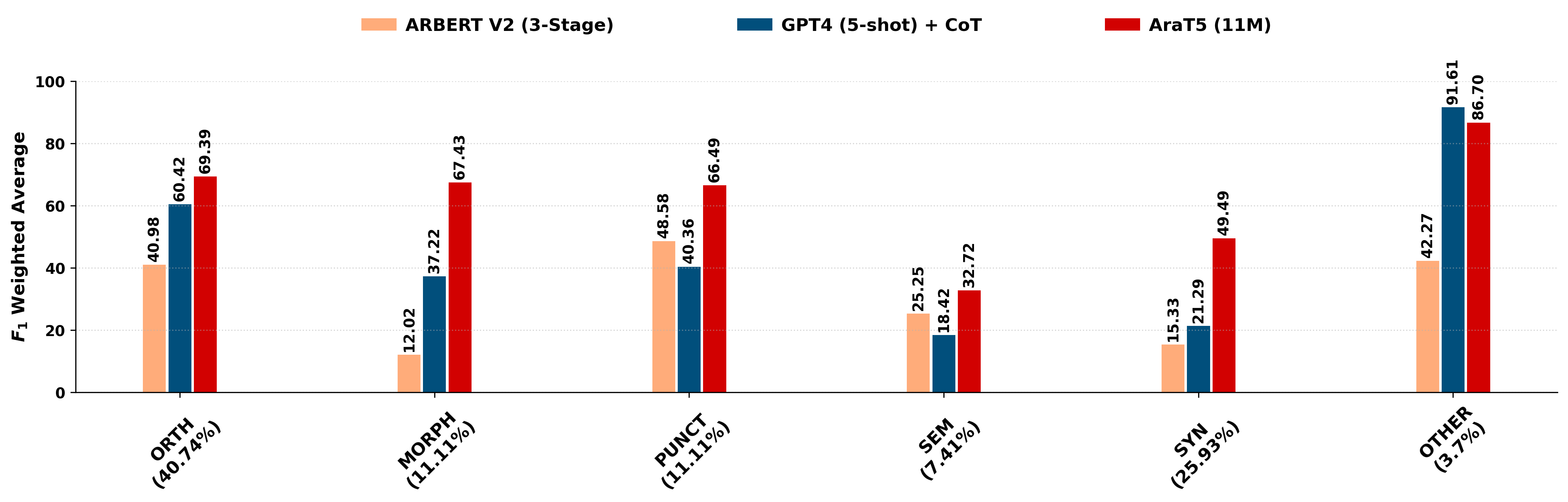}
\caption{F\textsubscript{1} scores for each fine-grained error type on the QALB-2014 test set. The percentages in parentheses indicate the proportion of each error type.}
\label{fig:areta_plot}
\end{figure*}

\noindent{\textbf{Type of errors.}} We concentrate on seven error types using ARETA: \textit{Orthographic}, \textit{Morphological}, \textit{Syntactic}, \textit{Semantic}, \textit{Punctuation}, and \textit{Merge} and \textit{Split} errors. Table~\ref{tab:errors_type_examples} presents examples of each error alongside their translations. Top-performing systems from each approach, including ARBERT v2 (3-step), GPT-4 (5-shot) + CoT, and AraT5 fully trained on the AGEC dataset~\cite{SOLYMAN2021303}, are analyzed in correcting these errors.
\begin{table}[t]
\centering
\renewcommand{\arraystretch}{1.5}
 \resizebox{\columnwidth}{!}{%
\begin{tabular}{lHrr}
\toprule
\textbf{Error Type}                   & Description& \multicolumn{1}{c}{\textbf{Incorrect Sentence}} & \multicolumn{1}{c}{\textbf{Correct Sentence}}   \\ \toprule
\multirow{2}{*}{\textbf{Orthographic} }  & \multirow{2}{*}{}                       

&  \< الفرس . > \textcolor{red}{ \< يرب > }    \< الرجل >  

&   \< الفرس . > \textcolor{green}{ \< يركب > }    \< الرجل >   

\\
   &                                          & \textit{The man \textcolor{red}{rears} the horse.}                         & \textit{The man \textcolor{green}{rides} the horse. }  \\ 
\multirow{2}{*}{\textbf{Punctuations}}  & \multirow{2}{*}{}                       
&  \<  يركب الفرس . >  \textcolor{red}{ \< ، > }    \< الرجل >  
& \<الرجل يركب الفرس .>  
    \\
                                      &                                          & The man\textcolor{red}{,} rides the horse.                      & The man rides the horse.                        \\

\multirow{2}{*}{\textbf{Syntax}}        & \multirow{2}{*}{}      
&  \textcolor{red}{ \< فرس .>  } \<وجد رجلا يركب>   
&  \textcolor{green}{ \< فرسا .>  } \<وجد رجلا يركب>   

\\
  &                                          & He found a man riding a \textcolor{red}{hors}.                  & He found a man riding a \textcolor{green}{horse}.                  \\
                                      
\multirow{2}{*}{\textbf{Merge}}         & \multirow{2}{*}{}                    

&  \< الفرس . >  \< سيركب > \textcolor{red}{ \< غداالرجل >  }
&  \< الفرس . >  \< سيركب > \textcolor{green}{ \< غدا الرجل >  }
\\

                                      &                                          & \textcolor{red}{Tomorrowtheman} will ride the horse.             & \textcolor{green}{Tomorrow the man} will ride the horse.           \\


\multirow{2}{*}{\textbf{Splits}}        & \multirow{2}{*}{}                     
&  \<  الفرس . >   \textcolor{red}{ \< ير كب > }    \< غدا الرجل >  
&  \<  الفرس . >   \textcolor{green}{ \< يركب > }    \< غدا الرجل >  

\\                                      &                                          & The man \textcolor{red}{ri des} the horse.                       & The man \textcolor{green}{rides} the horse.                        \\

\multirow{2}{*}{\textbf{Semantic}}      &                                         
&  \< ظهر الفرس . >   \textcolor{red}{ \< في  > }    \<  الرجل يجلس>  
&  \< ظهر الفرس . >   \textcolor{green}{ \< على  > }    \<  الرجل يجلس>  
\\
                                      &                                          & The man is sitting \textcolor{red}{in} the horse's back.         & The man is sitting \textcolor{green}{on} the horse's back.         \\


\multirow{2}{*}{\textbf{Morphological}} &                                  
&  \<  الفرس . >   \textcolor{red}{ \< ركب > }    \< غدا الرجل >  
&  \<  الفرس . >   \textcolor{green}{ \< سيركب > }    \< غدا الرجل >  

\\
                                      &                                          & Tomorrow the man \textcolor{red}{rode} the horse.                & Tomorrow the man \textcolor{green}{will ride} the horse.          \\

                                      \toprule
\end{tabular}}
\caption{Examples of Merge, Morphological, Orthographic, Punctuation, Semantic, Split, and Syntactic errors, along with their corresponding corrections and English translations.}\label{tab:errors_type_examples}
\end{table}

Figure~\ref{fig:areta_plot} illustrates the performance of each model under various error type categories. AraT5, fully trained on the AGEC dataset, surpasses all other models across all error    categories. In particular, it excels in handling \textit{Orthographic} (\text{ORTH}) errors, \textit{Morphological} (\text{MORPH}) errors, and \textit{Punctuation} (\text{PUNCT}) errors, consistently achieving over $65$ F\textsubscript{1} score. However, it is worth observing that all models encounter challenges with \textit{Semantic} (\text{SEM}) and \textit{Syntactic} (\text{SYN}) errors. These disparate outcomes underscore the significance of selecting the appropriate model based on the error types prevalent in a specific dataset.

\noindent{\subsection{Normalization Methods.}} 
\noindent In addition to the \textit{`Exact Match'} score, we also analyze system performance under different normalization methods. Looking at Table~\ref{tab: v2}, setting under `No punctuation' leads to an increase in scores across all models, underscoring the models' limitations in handling punctuation errors. Another noteworthy observation is the drop in F\textsubscript{1} scores when Alif/Ya errors are removed, illustrating the models' dependency on Alif/Ya features in making correction.
\begin{table*}[!htp]
\centering
\scriptsize
\renewcommand{\arraystretch}{1.2}
\resizebox{\textwidth}{!}{
\begin{tabular}{llrrrr|rrrr|rrrr|rrrrr}\toprule
\multirow{2}{*}{\textbf{Test Set}} &\multirow{2}{*}{\textbf{~~~~~~~~~~~~~~Models}} &\multicolumn{4}{c}{\textbf{Exact Match}} &\multicolumn{4}{c}{\textbf{No Alif / Ya Errors}} &\multicolumn{4}{c}{\textbf{No Punctuation}} &\multicolumn{4}{c}{\textbf{No Punctuation and Alif / Ya Errors}} \\\cmidrule{3-18}
& &\textbf{P} &\textbf{R} &\textbf{F\textsubscript{1.0}} &\textbf{F\textsubscript{0.5}} &\textbf{P} &\textbf{R} &\textbf{F\textsubscript{1.0}} &\textbf{F\textsubscript{0.5}} &\textbf{P} &\textbf{R} &\textbf{F\textsubscript{1.0}} &\textbf{F\textsubscript{0.5}} &\textbf{P} &\textbf{R} &\textbf{F\textsubscript{1.0}} &\textbf{F\textsubscript{0.5}} \\\midrule
\multirow{6}{*}{\textbf{QALB-2014}} 
&\citet{SOLYMAN2021303} &79.06 &65.79 &71.82 &75.99 &- &- &- &- &- &- &- &- &- &- &- &- \\
&AraT5 (11m) &\textbf{77.12} &\textbf{67.85} &\textbf{72.19} &\textbf{75.07} &\textbf{62.04} &\textbf{52.69} &\textbf{56.99 }&\textbf{59.91}&\textbf{86.57} &\textbf{82.70} &\textbf{84.59} &\textbf{85.77} &\textbf{79.32} &\textbf{67.19} &\textbf{72.75} &\textbf{76.56} \\
&GPT4 (5-shot) &69.46 &61.96 &65.49 &67.82 &58.44 &51.47 &54.73 &56.90 &74.59 &78.15 &76.33 &75.28 &60.06 &65.75 &62.78 &61.12 \\
&ARBERT V2 (3-step) &74.39 &47.62 &58.07 &66.87 &65.25 &41.58 &50.79 &58.58 &77.00 &46.00 &57.60 &67.85 &56.99 &28.90 &38.35 &47.71 \\
& \citet{mohit2014first} &73.34 &63.23 &67.91 &71.07 &64.05 &50.86 &56.7 &60.89 &76.99 &49.91 &60.56 &69.45 &76.99 &49.91 &60.56 &69.45 \\ \cmidrule{2-18}
\multirow{6}{*}{\textbf{QALB-2015}} 
&\citet{SOLYMAN2021303} &80.23 &63.59 &70.91 &76.24 &- &- &- &- &- &- &- &- &- &- &- &- \\
&AraT5 (11m) &\textbf{72.41} &\textbf{74.12} &\textbf{73.26} &\textbf{72.75} &\textbf{55.95} &\textbf{43.53} &\textbf{48.96} &\textbf{52.93} &\textbf{85.46} &\textbf{72.56} &\textbf{78.48} &\textbf{82.53} &\textbf{75.23} &\textbf{52.56} &\textbf{61.88} &\textbf{69.26} \\
&ChatGPT (3-shot) + EP &52.33 &47.57 &49.83 &54.10 &37.93 &39.97 &38.92 &32.95 &53.38 &56.63 &54.96 &54.00 &33.33 &46.77 &38.92 &35.36 \\
&ARBERT V2 (3-step) &74.20 &53.80 &62.37 &68.97 &57.30 &38.50 &46.06 &52.20 &66.70 &61.50 &63.99 &65.59 &71.24 &38.50 &49.99 &60.88 \\
&\citet{rozovskaya2015second}  &88.85 &61.76 &72.87 &81.68 &84.25 &43.29 &57.19 &70.84 &85.8 &77.98 &81.7 &84.11 &80.12 &58.24 &67.45 &74.52 \\\bottomrule
\end{tabular}}
\caption{Results on the Test sets of QALB-2014, QALB-2015 under Normalization Methods}\label{tab: v2}
\end{table*}

\section{Discussion}\label{D}

\noindent\textbf{LLMs and ChatGPT.} ChatGPT demonstrates a remarkable ability to outperform other fully trained models by learning from only a few examples, particularly five-shot under both few-shot CoT and EP prompting strategies. Nevertheless, ChatGPT's performance lags behind AraT5 and AraBART, suggesting potential areas for improvements in prompting strategies to fully exploit ChatGPT models. Larger models, such as Vicuna-13B, as well as those trained on multilingual datasets like  Bactrian-X$_{\textit{llama}}$-7B and  Bactrian-X$_{\textit{bloom}}$-7B, tend to perform better, with F\textsubscript{1} scores of $58.30$, $50.10$, and $52.50$ respectively. However, these models fail to match ChatGPT's performance in AGEC tasks. This reinforces ChatGPT's superiority in this domain.

\noindent\textbf{Data augmentation techniques.}
Data augmentation results underscore the potential of synthetic data, generated by ChatGPT, in enhancing model performance. Moreover, our findings reveal that not just the quantity, but the quality of synthetic data, is crucial for achieving optimal performance. The relative underperformance of models further trained with \textit{'out of domain'} data examples emphasizes this conclusion. Furthermore, our results on scaled datasets indicate a trade-off between precision and recall. As the size of the dataset increases, precision improves, while recall drops. This trend is apparent across all dataset sizes.

\noindent\textbf{Sequence tagging approach.}
These models exhibit high precision scores and relatively low recall scores, suggesting their strengths in making corrections rather than detecting errors. This trend can be explained by the absence of \textit{G-transformations}. For instance, in the case of English GECToR models, \textit{g-transformations} enable a variety of changes, such as case alterations and grammatical transformations. However, crafting effective \textit{G-transformations} for Arabic, a language with rich morphological features, poses significant challenges, limiting the model's ability to effectively detect errors.






\section{Conclusion}\label{C}
This paper provided a detailed exploration of the potential of LLMs, with a particular emphasis on ChatGPT for AGEC. Our study highlights ChatGPT's promising capabilities, in low-resource scenarios, as evidenced by its competitive performance on few-shot setttings. However, AraT5 and AraBART still exhibit superior results across various settings and error types. Our findings also emphasize the role of high-quality synthetic data, reinforcing that both quantity and quality matter in achieving optimal performance. Moreover, our work unveils trade-offs between precision and recall in relation to dataset size and throughout all the other experimental settings. These insight, again, could inform future strategies for improving GEC systems. Although our exploration of ChatGPT's performance on Arabic GEC tasks showcases encouraging results, it also uncovers areas ripe for further study. Notably, there remains significant room for improvement in GEC systems, particularly within the context of low-resource languages. Future research  may include refining prompting strategies, enhancing synthetic data generation techniques, and addressing the complexities and rich morphological features inherent in the Arabic language.



\bibliography{anthology,custom,citations}

\begin{thebibliography}{49}
\expandafter\ifx\csname natexlab\endcsname\relax\def\natexlab#1{#1}\fi

\bibitem[{Abdul-Mageed et~al.(2021)Abdul-Mageed, Elmadany, and
  Nagoudi}]{abdul-mageed-etal-2021-arbert}
Muhammad Abdul-Mageed, AbdelRahim Elmadany, and El~Moatez~Billah Nagoudi. 2021.
\newblock \href {https://doi.org/10.18653/v1/2021.acl-long.551} {{ARBERT} {\&}
  {MARBERT}: Deep bidirectional transformers for {A}rabic}.
\newblock In \emph{Proceedings of the 59th Annual Meeting of the Association
  for Computational Linguistics and the 11th International Joint Conference on
  Natural Language Processing (Volume 1: Long Papers)}, pages 7088--7105,
  Online. Association for Computational Linguistics.

\bibitem[{Abdul-Mageed et~al.(2020)Abdul-Mageed, Zhang, Bouamor, and
  Habash}]{abdul-mageed-etal-2020-nadi}
Muhammad Abdul-Mageed, Chiyu Zhang, Houda Bouamor, and Nizar Habash. 2020.
\newblock \href {https://aclanthology.org/2020.wanlp-1.9} {{NADI} 2020: The
  first nuanced {A}rabic dialect identification shared task}.
\newblock In \emph{Proceedings of the Fifth Arabic Natural Language Processing
  Workshop}, pages 97--110, Barcelona, Spain (Online). Association for
  Computational Linguistics.

\bibitem[{Alfaifi and Atwell(2012)}]{alfaifi2012arabic}
Abdullah Alfaifi and Eric Atwell. 2012.
\newblock Arabic learner corpora (alc): A taxonomy of coding errors.
\newblock In \emph{The 8th International Computing Conference in Arabic}.

\bibitem[{Awasthi et~al.(2019)Awasthi, Sarawagi, Goyal, Ghosh, and
  Piratla}]{awasthi-etal-2019-parallel}
Abhijeet Awasthi, Sunita Sarawagi, Rasna Goyal, Sabyasachi Ghosh, and Vihari
  Piratla. 2019.
\newblock \href {https://doi.org/10.18653/v1/D19-1435} {Parallel iterative edit
  models for local sequence transduction}.
\newblock In \emph{Proceedings of the 2019 Conference on Empirical Methods in
  Natural Language Processing and the 9th International Joint Conference on
  Natural Language Processing (EMNLP-IJCNLP)}, pages 4260--4270, Hong Kong,
  China. Association for Computational Linguistics.

\bibitem[{Belkebir and Habash(2021)}]{belkebir2021automatic}
Riadh Belkebir and Nizar Habash. 2021.
\newblock Automatic error type annotation for arabic.
\newblock \emph{arXiv preprint arXiv:2109.08068}.

\bibitem[{Brown et~al.(2020)Brown, Mann, Ryder, Subbiah, Kaplan, Dhariwal,
  Neelakantan, Shyam, Sastry, Askell et~al.}]{brown2020language}
Tom Brown, Benjamin Mann, Nick Ryder, Melanie Subbiah, Jared~D Kaplan, Prafulla
  Dhariwal, Arvind Neelakantan, Pranav Shyam, Girish Sastry, Amanda Askell,
  et~al. 2020.
\newblock Language models are few-shot learners.
\newblock \emph{Advances in neural information processing systems},
  33:1877--1901.

\bibitem[{Bryant et~al.(2022)Bryant, Yuan, Qorib, Cao, Ng, and
  Briscoe}]{bryant2022grammatical}
Christopher Bryant, Zheng Yuan, Muhammad~Reza Qorib, Hannan Cao, Hwee~Tou Ng,
  and Ted Briscoe. 2022.
\newblock Grammatical error correction: A survey of the state of the art.
\newblock \emph{arXiv preprint arXiv:2211.05166}.

\bibitem[{Chiang et~al.(2023)Chiang, Li, Lin, Sheng, Wu, Zhang, Zheng, Zhuang,
  Zhuang, Gonzalez, Stoica, and Xing}]{vicuna2023}
Wei-Lin Chiang, Zhuohan Li, Zi~Lin, Ying Sheng, Zhanghao Wu, Hao Zhang, Lianmin
  Zheng, Siyuan Zhuang, Yonghao Zhuang, Joseph~E. Gonzalez, Ion Stoica, and
  Eric~P. Xing. 2023.
\newblock \href {https://lmsys.org/blog/2023-03-30-vicuna/} {Vicuna: An
  open-source chatbot impressing gpt-4 with 90\%* chatgpt quality}.

\bibitem[{Chollampatt and Ng(2018)}]{Chollampatt_Ng_2018}
Shamil Chollampatt and Hwee~Tou Ng. 2018.
\newblock \href {https://doi.org/10.1609/aaai.v32i1.12069} {A multilayer
  convolutional encoder-decoder neural network for grammatical error
  correction}.
\newblock \emph{Proceedings of the AAAI Conference on Artificial Intelligence},
  32(1).

\bibitem[{Chung et~al.(2022)Chung, Hou, Longpre, Zoph, Tay, Fedus, Li, Wang,
  Dehghani, Brahma et~al.}]{chung2022scaling}
Hyung~Won Chung, Le~Hou, Shayne Longpre, Barret Zoph, Yi~Tay, William Fedus,
  Eric Li, Xuezhi Wang, Mostafa Dehghani, Siddhartha Brahma, et~al. 2022.
\newblock Scaling instruction-finetuned language models.
\newblock \emph{arXiv preprint arXiv:2210.11416}.

\bibitem[{Costa-juss{\`a} et~al.(2022)Costa-juss{\`a}, Cross, {\c{C}}elebi,
  Elbayad, Heafield, Heffernan, Kalbassi, Lam, Licht, Maillard
  et~al.}]{costa2022no}
Marta~R Costa-juss{\`a}, James Cross, Onur {\c{C}}elebi, Maha Elbayad, Kenneth
  Heafield, Kevin Heffernan, Elahe Kalbassi, Janice Lam, Daniel Licht, Jean
  Maillard, et~al. 2022.
\newblock No language left behind: Scaling human-centered machine translation.
\newblock \emph{arXiv preprint arXiv:2207.04672}.

\bibitem[{Dahlmeier and Ng(2012)}]{dahlmeier-ng-2012-better}
Daniel Dahlmeier and Hwee~Tou Ng. 2012.
\newblock \href {https://aclanthology.org/N12-1067} {Better evaluation for
  grammatical error correction}.
\newblock In \emph{Proceedings of the 2012 Conference of the North {A}merican
  Chapter of the Association for Computational Linguistics: Human Language
  Technologies}, pages 568--572, Montr{\'e}al, Canada. Association for
  Computational Linguistics.

\bibitem[{Eddine et~al.(2022)Eddine, Tomeh, Habash, Roux, and
  Vazirgiannis}]{eddine2022arabart}
Moussa~Kamal Eddine, Nadi Tomeh, Nizar Habash, Joseph~Le Roux, and Michalis
  Vazirgiannis. 2022.
\newblock Arabart: a pretrained arabic sequence-to-sequence model for
  abstractive summarization.
\newblock \emph{arXiv preprint arXiv:2203.10945}.

\bibitem[{Elmadany et~al.(2022)Elmadany, Nagoudi, and
  Abdul-Mageed}]{elmadany2022orca}
AbdelRahim Elmadany, El~Moatez~Billah Nagoudi, and Muhammad Abdul-Mageed. 2022.
\newblock Orca: A challenging benchmark for arabic language understanding.
\newblock \emph{arXiv preprint arXiv:2212.10758}.

\bibitem[{Fang et~al.(2023)Fang, Yang, Lan, Wong, Hu, Chao, and
  Zhang}]{fang2023chatgpt}
Tao Fang, Shu Yang, Kaixin Lan, Derek~F Wong, Jinpeng Hu, Lidia~S Chao, and Yue
  Zhang. 2023.
\newblock Is chatgpt a highly fluent grammatical error correction system? a
  comprehensive evaluation.
\newblock \emph{arXiv preprint arXiv:2304.01746}.

\bibitem[{Felice et~al.(2014)Felice, Yuan, Andersen, Yannakoudakis, and
  Kochmar}]{felice-etal-2014-grammatical}
Mariano Felice, Zheng Yuan, {\O}istein~E. Andersen, Helen Yannakoudakis, and
  Ekaterina Kochmar. 2014.
\newblock \href {https://doi.org/10.3115/v1/W14-1702} {Grammatical error
  correction using hybrid systems and type filtering}.
\newblock In \emph{Proceedings of the Eighteenth Conference on Computational
  Natural Language Learning: Shared Task}, pages 15--24, Baltimore, Maryland.
  Association for Computational Linguistics.

\bibitem[{Flachs et~al.(2021)Flachs, Stahlberg, and
  Kumar}]{flachs-etal-2021-data}
Simon Flachs, Felix Stahlberg, and Shankar Kumar. 2021.
\newblock \href {https://aclanthology.org/2021.bea-1.12} {Data strategies for
  low-resource grammatical error correction}.
\newblock In \emph{Proceedings of the 16th Workshop on Innovative Use of NLP
  for Building Educational Applications}, pages 117--122, Online. Association
  for Computational Linguistics.

\bibitem[{Gong et~al.(2022)Gong, Liu, Huang, and
  Zhang}]{gong-etal-2022-revisiting}
Peiyuan Gong, Xuebo Liu, Heyan Huang, and Min Zhang. 2022.
\newblock \href {https://aclanthology.org/2022.emnlp-main.463} {Revisiting
  grammatical error correction evaluation and beyond}.
\newblock In \emph{Proceedings of the 2022 Conference on Empirical Methods in
  Natural Language Processing}, pages 6891--6902, Abu Dhabi, United Arab
  Emirates. Association for Computational Linguistics.

\bibitem[{Grundkiewicz et~al.(2019)Grundkiewicz, Junczys-Dowmunt, and
  Heafield}]{grundkiewicz-etal-2019-neural}
Roman Grundkiewicz, Marcin Junczys-Dowmunt, and Kenneth Heafield. 2019.
\newblock \href {https://doi.org/10.18653/v1/W19-4427} {Neural grammatical
  error correction systems with unsupervised pre-training on synthetic data}.
\newblock In \emph{Proceedings of the Fourteenth Workshop on Innovative Use of
  NLP for Building Educational Applications}, pages 252--263, Florence, Italy.
  Association for Computational Linguistics.

\bibitem[{Habash and Palfreyman(2022)}]{habash-palfreyman-2022-zaebuc}
Nizar Habash and David Palfreyman. 2022.
\newblock \href {https://aclanthology.org/2022.lrec-1.9} {{ZAEBUC}: An
  annotated {A}rabic-{E}nglish bilingual writer corpus}.
\newblock In \emph{Proceedings of the Thirteenth Language Resources and
  Evaluation Conference}, pages 79--88, Marseille, France. European Language
  Resources Association.

\bibitem[{Junczys-Dowmunt et~al.(2018{\natexlab{a}})Junczys-Dowmunt,
  Grundkiewicz, Guha, and Heafield}]{junczys-dowmunt-etal-2018-approaching}
Marcin Junczys-Dowmunt, Roman Grundkiewicz, Shubha Guha, and Kenneth Heafield.
  2018{\natexlab{a}}.
\newblock \href {https://doi.org/10.18653/v1/N18-1055} {Approaching neural
  grammatical error correction as a low-resource machine translation task}.
\newblock In \emph{Proceedings of the 2018 Conference of the North {A}merican
  Chapter of the Association for Computational Linguistics: Human Language
  Technologies, Volume 1 (Long Papers)}, pages 595--606, New Orleans,
  Louisiana. Association for Computational Linguistics.

\bibitem[{Junczys-Dowmunt et~al.(2018{\natexlab{b}})Junczys-Dowmunt,
  Grundkiewicz, Guha, and Heafield}]{junczys2018approaching}
Marcin Junczys-Dowmunt, Roman Grundkiewicz, Shubha Guha, and Kenneth Heafield.
  2018{\natexlab{b}}.
\newblock Approaching neural grammatical error correction as a low-resource
  machine translation task.
\newblock \emph{arXiv preprint arXiv:1804.05940}.

\bibitem[{Kojima et~al.(2022)Kojima, Gu, Reid, Matsuo, and
  Iwasawa}]{kojima2022large}
Takeshi Kojima, Shixiang~Shane Gu, Machel Reid, Yutaka Matsuo, and Yusuke
  Iwasawa. 2022.
\newblock Large language models are zero-shot reasoners.
\newblock \emph{arXiv preprint arXiv:2205.11916}.

\bibitem[{Li et~al.(2023)Li, Koto, Wu, Aji, and Baldwin}]{bactrian}
Haonan Li, Fajri Koto, Minghao Wu, Alham~Fikri Aji, and Timothy Baldwin. 2023.
\newblock Bactrian-x: A multilingual replicable instruction-following model.
\newblock \url{https://github.com/MBZUAI-nlp/Bactrian-X}.

\bibitem[{Malmi et~al.(2019)Malmi, Krause, Rothe, Mirylenka, and
  Severyn}]{malmi2019encode}
Eric Malmi, Sebastian Krause, Sascha Rothe, Daniil Mirylenka, and Aliaksei
  Severyn. 2019.
\newblock Encode, tag, realize: High-precision text editing.
\newblock \emph{arXiv preprint arXiv:1909.01187}.

\bibitem[{Mohit et~al.(2014)Mohit, Rozovskaya, Habash, Zaghouani, and
  Obeid}]{mohit2014first}
Behrang Mohit, Alla Rozovskaya, Nizar Habash, Wajdi Zaghouani, and Ossama
  Obeid. 2014.
\newblock The first qalb shared task on automatic text correction for arabic.
\newblock In \emph{Proceedings of the EMNLP 2014 Workshop on Arabic Natural
  Language Processing (ANLP)}, pages 39--47.

\bibitem[{Muennighoff et~al.(2022)Muennighoff, Wang, Sutawika, Roberts,
  Biderman, Scao, Bari, Shen, Yong, Schoelkopf, Tang, Radev, Aji, Almubarak,
  Albanie, Alyafeai, Webson, Raff, and Raffel}]{muennighoff2022crosslingual}
Niklas Muennighoff, Thomas Wang, Lintang Sutawika, Adam Roberts, Stella
  Biderman, Teven~Le Scao, M~Saiful Bari, Sheng Shen, Zheng-Xin Yong, Hailey
  Schoelkopf, Xiangru Tang, Dragomir Radev, Alham~Fikri Aji, Khalid Almubarak,
  Samuel Albanie, Zaid Alyafeai, Albert Webson, Edward Raff, and Colin Raffel.
  2022.
\newblock \href {http://arxiv.org/abs/2211.01786} {Crosslingual generalization
  through multitask finetuning}.

\bibitem[{Nagoudi et~al.(2021)Nagoudi, Elmadany, and
  Abdul-Mageed}]{nagoudi2021arat5}
El~Moatez~Billah Nagoudi, AbdelRahim Elmadany, and Muhammad Abdul-Mageed. 2021.
\newblock Arat5: Text-to-text transformers for arabic language generation.
\newblock \emph{arXiv preprint arXiv:2109.12068}.

\bibitem[{Ng et~al.(2014)Ng, Wu, Briscoe, Hadiwinoto, Susanto, and
  Bryant}]{ng-etal-2014-conll}
Hwee~Tou Ng, Siew~Mei Wu, Ted Briscoe, Christian Hadiwinoto, Raymond~Hendy
  Susanto, and Christopher Bryant. 2014.
\newblock \href {https://doi.org/10.3115/v1/W14-1701} {The {C}o{NLL}-2014
  shared task on grammatical error correction}.
\newblock In \emph{Proceedings of the Eighteenth Conference on Computational
  Natural Language Learning: Shared Task}, pages 1--14, Baltimore, Maryland.
  Association for Computational Linguistics.

\bibitem[{Ng et~al.(2013)Ng, Wu, Wu, Hadiwinoto, and
  Tetreault}]{ng-etal-2013-conll}
Hwee~Tou Ng, Siew~Mei Wu, Yuanbin Wu, Christian Hadiwinoto, and Joel Tetreault.
  2013.
\newblock \href {https://aclanthology.org/W13-3601} {The {C}o{NLL}-2013 shared
  task on grammatical error correction}.
\newblock In \emph{Proceedings of the Seventeenth Conference on Computational
  Natural Language Learning: Shared Task}, pages 1--12, Sofia, Bulgaria.
  Association for Computational Linguistics.

\bibitem[{Omelianchuk et~al.(2020)Omelianchuk, Atrasevych, Chernodub, and
  Skurzhanskyi}]{omelianchuk2020gector}
Kostiantyn Omelianchuk, Vitaliy Atrasevych, Artem Chernodub, and Oleksandr
  Skurzhanskyi. 2020.
\newblock Gector--grammatical error correction: tag, not rewrite.
\newblock \emph{arXiv preprint arXiv:2005.12592}.

\bibitem[{Ouyang et~al.(2022)Ouyang, Wu, Jiang, Almeida, Wainwright, Mishkin,
  Zhang, Agarwal, Slama, Ray et~al.}]{ouyang2022training}
Long Ouyang, Jeffrey Wu, Xu~Jiang, Diogo Almeida, Carroll Wainwright, Pamela
  Mishkin, Chong Zhang, Sandhini Agarwal, Katarina Slama, Alex Ray, et~al.
  2022.
\newblock Training language models to follow instructions with human feedback.
\newblock \emph{Advances in Neural Information Processing Systems},
  35:27730--27744.

\bibitem[{Raffel et~al.(2020)Raffel, Shazeer, Roberts, Lee, Narang, Matena,
  Zhou, Li, and Liu}]{raffel2020exploring}
Colin Raffel, Noam Shazeer, Adam Roberts, Katherine Lee, Sharan Narang, Michael
  Matena, Yanqi Zhou, Wei Li, and Peter~J Liu. 2020.
\newblock Exploring the limits of transfer learning with a unified text-to-text
  transformer.
\newblock \emph{The Journal of Machine Learning Research}, 21(1):5485--5551.

\bibitem[{Rothe et~al.(2021)Rothe, Mallinson, Malmi, Krause, and
  Severyn}]{rothe-etal-2021-simple}
Sascha Rothe, Jonathan Mallinson, Eric Malmi, Sebastian Krause, and Aliaksei
  Severyn. 2021.
\newblock \href {https://doi.org/10.18653/v1/2021.acl-short.89} {A simple
  recipe for multilingual grammatical error correction}.
\newblock In \emph{Proceedings of the 59th Annual Meeting of the Association
  for Computational Linguistics and the 11th International Joint Conference on
  Natural Language Processing (Volume 2: Short Papers)}, pages 702--707,
  Online. Association for Computational Linguistics.

\bibitem[{Rozovskaya et~al.(2015{\natexlab{a}})Rozovskaya, Bouamor, Habash,
  Zaghouani, Obeid, and Mohit}]{rozovskaya2015second}
Alla Rozovskaya, Houda Bouamor, Nizar Habash, Wajdi Zaghouani, Ossama Obeid,
  and Behrang Mohit. 2015{\natexlab{a}}.
\newblock The second qalb shared task on automatic text correction for arabic.
\newblock In \emph{Proceedings of the Second workshop on Arabic natural
  language processing}, pages 26--35.

\bibitem[{Rozovskaya et~al.(2015{\natexlab{b}})Rozovskaya, Bouamor, Habash,
  Zaghouani, Obeid, and Mohit}]{rozovskaya-etal-2015-second}
Alla Rozovskaya, Houda Bouamor, Nizar Habash, Wajdi Zaghouani, Ossama Obeid,
  and Behrang Mohit. 2015{\natexlab{b}}.
\newblock \href {https://doi.org/10.18653/v1/W15-3204} {The second {QALB}
  shared task on automatic text correction for {A}rabic}.
\newblock In \emph{Proceedings of the Second Workshop on {A}rabic Natural
  Language Processing}, pages 26--35, Beijing, China. Association for
  Computational Linguistics.

\bibitem[{Sanh et~al.(2021)Sanh, Webson, Raffel, Bach, Sutawika, Alyafeai,
  Chaffin, Stiegler, Scao, Raja et~al.}]{sanh2021multitask}
Victor Sanh, Albert Webson, Colin Raffel, Stephen~H Bach, Lintang Sutawika,
  Zaid Alyafeai, Antoine Chaffin, Arnaud Stiegler, Teven~Le Scao, Arun Raja,
  et~al. 2021.
\newblock Multitask prompted training enables zero-shot task generalization.
\newblock \emph{arXiv preprint arXiv:2110.08207}.

\bibitem[{Solyman et~al.(2022)Solyman, Wang, Tao, Elhag, Zhang, and
  Mahmoud}]{solyman2022automatic}
Aiman Solyman, Zhenyu Wang, Qian Tao, Arafat Abdulgader~Mohammed Elhag, Rui
  Zhang, and Zeinab Mahmoud. 2022.
\newblock Automatic arabic grammatical error correction based on
  expectation-maximization routing and target-bidirectional agreement.
\newblock \emph{Knowledge-Based Systems}, 241:108180.

\bibitem[{Solyman et~al.(2023)Solyman, Zappatore, Zhenyu, Mahmoud, Alfatemi,
  Ibrahim, and Gabralla}]{SOLYMAN2023101572}
Aiman Solyman, Marco Zappatore, Wang Zhenyu, Zeinab Mahmoud, Ali Alfatemi,
  Ashraf~Osman Ibrahim, and Lubna~Abdelkareim Gabralla. 2023.
\newblock \href {https://doi.org/https://doi.org/10.1016/j.jksuci.2023.101572}
  {Optimizing the impact of data augmentation for low-resource grammatical
  error correction}.
\newblock \emph{Journal of King Saud University - Computer and Information
  Sciences}, 35(6):101572.

\bibitem[{Solyman et~al.(2021)Solyman, Zhenyu, Qian, Elhag, Toseef, and
  Aleibeid}]{SOLYMAN2021303}
Aiman Solyman, Wang Zhenyu, Tao Qian, Arafat Abdulgader~Mohammed Elhag,
  Muhammad Toseef, and Zeinab Aleibeid. 2021.
\newblock \href {https://doi.org/https://doi.org/10.1016/j.eij.2020.12.001}
  {Synthetic data with neural machine translation for automatic correction in
  arabic grammar}.
\newblock \emph{Egyptian Informatics Journal}, 22(3):303--315.

\bibitem[{Taori et~al.(2023)Taori, Gulrajani, Zhang, Dubois, Li, Guestrin,
  Liang, and Hashimoto}]{alpaca}
Rohan Taori, Ishaan Gulrajani, Tianyi Zhang, Yann Dubois, Xuechen Li, Carlos
  Guestrin, Percy Liang, and Tatsunori~B. Hashimoto. 2023.
\newblock Stanford alpaca: An instruction-following llama model.
\newblock \url{https://github.com/tatsu-lab/stanford_alpaca}.

\bibitem[{Tarnavskyi et~al.(2022)Tarnavskyi, Chernodub, and
  Omelianchuk}]{tarnavskyi-etal-2022-ensembling}
Maksym Tarnavskyi, Artem Chernodub, and Kostiantyn Omelianchuk. 2022.
\newblock \href {https://doi.org/10.18653/v1/2022.acl-long.266} {Ensembling and
  knowledge distilling of large sequence taggers for grammatical error
  correction}.
\newblock In \emph{Proceedings of the 60th Annual Meeting of the Association
  for Computational Linguistics (Volume 1: Long Papers)}, pages 3842--3852,
  Dublin, Ireland. Association for Computational Linguistics.

\bibitem[{Touvron et~al.(2023)Touvron, Lavril, Izacard, Martinet, Lachaux,
  Lacroix, Rozi{\`e}re, Goyal, Hambro, Azhar et~al.}]{touvron2023llama}
Hugo Touvron, Thibaut Lavril, Gautier Izacard, Xavier Martinet, Marie-Anne
  Lachaux, Timoth{\'e}e Lacroix, Baptiste Rozi{\`e}re, Naman Goyal, Eric
  Hambro, Faisal Azhar, et~al. 2023.
\newblock Llama: Open and efficient foundation language models.
\newblock \emph{arXiv preprint arXiv:2302.13971}.

\bibitem[{Watson et~al.(2018)Watson, Zalmout, and Habash}]{watson2018utilizing}
Daniel Watson, Nasser Zalmout, and Nizar Habash. 2018.
\newblock Utilizing character and word embeddings for text normalization with
  sequence-to-sequence models.
\newblock \emph{arXiv preprint arXiv:1809.01534}.

\bibitem[{Wei et~al.(2022)Wei, Wang, Schuurmans, Bosma, Chi, Le, and
  Zhou}]{wei2022chain}
Jason Wei, Xuezhi Wang, Dale Schuurmans, Maarten Bosma, Ed~Chi, Quoc Le, and
  Denny Zhou. 2022.
\newblock Chain of thought prompting elicits reasoning in large language
  models.
\newblock \emph{arXiv preprint arXiv:2201.11903}.

\bibitem[{Wu et~al.(2023)Wu, Wang, Wan, Jiao, and Lyu}]{wu2023chatgpt}
Haoran Wu, Wenxuan Wang, Yuxuan Wan, Wenxiang Jiao, and Michael Lyu. 2023.
\newblock Chatgpt or grammarly? evaluating chatgpt on grammatical error
  correction benchmark.
\newblock \emph{arXiv preprint arXiv:2303.13648}.

\bibitem[{Xie et~al.(2018)Xie, Genthial, Xie, Ng, and
  Jurafsky}]{xie-etal-2018-noising}
Ziang Xie, Guillaume Genthial, Stanley Xie, Andrew Ng, and Dan Jurafsky. 2018.
\newblock \href {https://doi.org/10.18653/v1/N18-1057} {Noising and denoising
  natural language: Diverse backtranslation for grammar correction}.
\newblock In \emph{Proceedings of the 2018 Conference of the North {A}merican
  Chapter of the Association for Computational Linguistics: Human Language
  Technologies, Volume 1 (Long Papers)}, pages 619--628, New Orleans,
  Louisiana. Association for Computational Linguistics.

\bibitem[{Xu et~al.(2023)Xu, Yang, Lin, Wang, Zhou, Zhang, and
  Mao}]{xu2023expertprompting}
Benfeng Xu, An~Yang, Junyang Lin, Quan Wang, Chang Zhou, Yongdong Zhang, and
  Zhendong Mao. 2023.
\newblock Expertprompting: Instructing large language models to be
  distinguished experts.
\newblock \emph{arXiv preprint arXiv:2305.14688}.

\bibitem[{Xue et~al.(2020)Xue, Constant, Roberts, Kale, Al-Rfou, Siddhant,
  Barua, and Raffel}]{xue2020mt5}
Linting Xue, Noah Constant, Adam Roberts, Mihir Kale, Rami Al-Rfou, Aditya
  Siddhant, Aditya Barua, and Colin Raffel. 2020.
\newblock mt5: A massively multilingual pre-trained text-to-text transformer.
\newblock \emph{arXiv preprint arXiv:2010.11934}.

\end{thebibliography}
\bibliographystyle{acl_natbib}
\newpage
\appendix
\clearpage


\section{Related Works} \label{app:RW_approaches}

\paragraph{Sequence to Sequence Approach.} Transformer-based Language Models (LMs) have been integral to advancements in GEC. These models have substantially transformed the perception of GEC, reframing it as a MT task. In this framework, erroneous sentences are considered as the source language, and the corrected versions as the target language. This perspective, which has led to SOTA results in the CONLL 2013 and 2014 shared tasks \cite{bryant2022grammatical,ng-etal-2013-conll,ng-etal-2014-conll}, reinterprets GEC as a low-resource or mid-resource MT task. Building on this paradigm, \citet{junczys-dowmunt-etal-2018-approaching} successfully adopted techniques from low-resource NMT and Statistical Machine Translation (SMT)-based GEC methods, leading to considerable improvements on both the CONLL and JFLEG datasets.

\paragraph{Sequence Tagging Approach.} Sequence tagging methods, another successful route to GEC, are showcased by models like GECToR \cite{omelianchuk2020gector}, LaserTagger \cite{malmi2019encode}, and the Parallel Iterative Edit (PIE) model ~\cite{awasthi-etal-2019-parallel}. By viewing GEC as a text editing task, these models make edits predictions instead of tokens, label sequences rather than generating them, and iteratively refine predictions to tackle dependencies. Employing a limited set of output tags, these models apply edit operations on the input sequence, reconstructing the output. This technique not only capably mirrors a significant chunk of the target training data, but it also diminishes the vocabulary size and establishes the output length as the source text's word count. Consequently, it curtails the number of training examples necessary for model accuracy, which is particularly beneficial in settings with sparse human-labeled data ~\cite{awasthi-etal-2019-parallel}.

\paragraph{Instruction Finetuning.} LLMs have revolutionized NLP, their vast data-learning capability enabling diverse task generalizations. Key to their enhancement has been instructional finetuning, which fortifies the model's directive response and mitigates the need for few-shot examples ~\cite{ouyang2022training,wei2022chain,sanh2021multitask}. A novel approach, Chain of Thought (CoT), directs LLMs through a series of natural language reasoning, generating superior outputs. Proven beneficial in 'Let’s think step by step' prompts ~\cite{wei2022chain}, CoT has harnessed LLMs for multi-task cognitive tasks ~\cite{kojima2022large} and achieved SOTA results in complex system-2 tasks like arithmetic and symbolic reasoning.

\paragraph{ChatGPT.} In the specific realm of GEC, LLMs have demonstrated its potential.  \citet{fang2023chatgpt} applied zero-shot and few-shot CoT settings using in-context learning for ChatGPT ~\cite{brown2020language} and evaluated its performance on three document-level English GEC test sets. Similarly, ~\citet{wu2023chatgpt} carried out an empirical study to assess the effectiveness of ChatGPT in GEC, in the CoNLL2014 benchmark dataset.

\section{Normalisation Table}\label{errors_normaliszation}
\begin{table*}[t]
\centering
\renewcommand{\arraystretch}{1.5}
 \resizebox{1.8\columnwidth}{!}{%

\begin{tabular}{lr}

\toprule
\textbf{Normalisation Method}             & \multicolumn{1}{c}{\textbf{Example}}                                                                                                                                                                                                               \\ \toprule
\textbf{Normal}                 &

\RL{نحن معشر العرب نعرف إلا الشماتة ، ولكن يجب أن ندرس هذه الحالة ونحن المخرج منها من الاقتصاد الإسلامي.}

\\
\textbf{No Alif/Ya}              & 
\RL{ نحن معشر العرب نعرف الا الشماتة ، ولكن يجب ان ندرس هذه الحالة ونحن المخرج منها من الاقتصاد الاسلامي.} 
\\

\textbf{No Punct}                &

\RL{نحن معشر العرب نعرف إلا الشماتة ولكن يجب أن ندرس هذه الحالة ونحن المخرج منها من الاقتصاد الإسلامي} 

\\
\textbf{No Alif/Ya \& Punct}  & 
\RL{نحن معشر العرب نعرف الا الشماتة ولكن يجب ان ندرس هذه الحالة ونحن المخرج منها من الاقتصاد الاسلامي}
\\ \toprule
\end{tabular}
}
\caption{Examples of normalized text: with Alif/Ya errors removed, punctuation removed, and both Alif/Ya errors and punctuation removed.  \label{tab:errors_normaliszation}}
\end{table*}


\section{Instructions for LLaMa}\label{app:IF}
\begin{table*}[h]
\centering
\renewcommand{\arraystretch}{1.95}
\resizebox{0.95\textwidth}{!}{
\begin{tabular}{lr}\toprule
\textbf{Translated in English} &\textbf{Instructions Samples} \\\midrule
Correct all written errors in the following text except for a thousand, ya and punctuation: &\RL{قم بتصحيح كل الأخطاء الكتابية في النص التالي ماعدا المتعلقة بالألف والياء وعلامات الترقيم:} \\
Please verify spelling, grammatical scrutiny, and correct all errors in the following sentence, except for punctuation: &\RL{الرجاء التدقيق الإملائي والتدقيق النحوي و تصحيح كل الأخطاء في الجملة التالية إلا الخاصة بعلامات الترقيم:} \\
Explore the grammatical errors and repair them except for punctuation marks such as a comma, or a question marks, etc: &\RL{قم بإستكشاف أخطاء التدقيق الإملائي وإصلاحها ماعدا المتعلقة بعلامات الترقيم كالفاصلة  أو علامة إستفهام ، إلخ:} \\
Can you correct all errors in the following text except those related to punctuation such as commas, periods, etc: &\RL{هل يمكنك كل الأخطاء الموجودة في النص التالي ماعدا المتعلقة بعلامات الترقيم كالفاصلة ، النقطة ، إلخ : } \\
Can you fix all spelling and grammatical errors, except for the mistakes of the "Alif" and "Ya": &\RL{هل يمكنك إصلاح كل الأخطاء الإملائية والنحوية ماعدا الأخطاء الخاصة بالألف والياء:} \\
Please explore the grammatical spelling errors and repair them all, except for the mistakes related to the "Alif" and "Ya" &\RL{الرجاء إستكشاف أخطاء التدقيق الإملائي النحوي وإصلاحها كلها ماعدا الأخطاء المتعلقة بالألف والياء:} \\
Correct all the written errors in the following text except for the "Alif" and "Ya": &\RL{قم بتصحيح كل الأخطاء الكتابية في النص التالي ماعدا المتعلقة بالألف والياء:} \\
Please correct all errors in the following sentence: &\RL{الرجاء تصحيح كل الأخطاء الموجودة في الجملة التالية:} \\
\bottomrule
\end{tabular}}
\caption{Different instructions used for instruction fine-tuning.}\label{tab:llamains}
\end{table*}
\begin{table}[]
\centering
\scriptsize
\resizebox{0.90\columnwidth}{!}{
\begin{tabular}{lr}
\toprule
~~~~~~~~~ \textbf{Fine-tune Instruction Example }\\
\toprule
\RL{ فيما يلي أمر توجيه يصف مهمة مرتبطة بمدخل لتزويد} \\
~~\RL{  النص بسياق اضافي. يرجى صياغة ردود مناسبة لتحقق} \\
~~~~~~~~~~~~~~~~~~~~~~~~~~~~~~~~~\RL{  الطلب بطريقة مناسبة و دقيقة.} \\
\colorbox{blue!25}{\#\#\# \RL{: الأمر/ التوجيه} }\\
\textbf{\textit{\RL{قم بتصحيح كل الأخطاء الكتابية في النص التالي:}}}\\
\colorbox{pink!55}{\#\#\#\RL{ : المدخل}} \\
\textbf{\textit{\< الفرس . > \textcolor{red}{ \< يرب > }    \< الرجل >}} \\
\colorbox{yellow!50}{\#\#\#\RL{ : الرد}} \\
\textbf{\textit{ \<  الفرس . > \textcolor{green}{  \< يركب > }    \< الرجل >}} \\
\bottomrule
\end{tabular}
}
\caption{Modified data format for the LLaMA instruction fine-tuning step.}\label{tab:llamadata}
\end{table}





\section{ALC Error Type Taxonomy }\label{app:acl}
\begin{table*}[t]
\centering
\renewcommand{\arraystretch}{1.}
\resizebox{0.85\textwidth}{!}{%
\begin{tabular}{ccl}
\toprule
\textbf{Class}                          & \textbf{Sub-class} & \textbf{Description}                                \\\toprule
\multirow{12}{*}{\textbf{Orthographic}} & \textbf{OH}        & \textbf{Hamza error}                                \\
                                        & \textbf{OT}        & Confusion in Ha and Ta Mutadarrifatin               \\
                                        & \textbf{OA}        & \textbf{Confusuion in Alif and Ya Mutadarrifatin}   \\
                                        & \textbf{OW}        & Confusion in Alif Fariqa                            \\
                                        & \textbf{ON}        & Confusion Between Nun and Tanwin                    \\
                                        & \textbf{OS}        & Shortening the long vowels                          \\
                                        & \textbf{OG}        & Lengthening the short vowels                        \\
                                        & \textbf{OC}        & Wrong order of word characters                      \\
                                        & \textbf{OR}        & Replacement in word character(s)                    \\
                                        & \textbf{OD}        & Additional character(s)                             \\
                                        & \textbf{OM}        & Missing character(s)                                \\
                                        & \textbf{OO}        & Other orthographic errors                           \\ \midrule
\multirow{9}{*}{\textbf{Morphological}} & \textbf{MI}        & Word inflection                                     \\
                                        & \textbf{MT}        & Verb tense                                          \\
                                        & \textbf{MO}        & Other morphological errors                          \\
                                        & \textbf{XF}        & Definiteness                                        \\
                                        & \textbf{XG}        & Gender                                              \\
                                        & \textbf{XN}        & Number                                              \\
                                        & \textbf{XT}        & Unnecessary word                                    \\
                                        & \textbf{XM}        & Missing word                                        \\
                                        & \textbf{XO}        & Other syntactic errors                              \\ \midrule
\multirow{3}{*}{\textbf{Semantic}}      & \textbf{SW}        & Word selection error                                \\
                                        & \textbf{SF}        & Fasl wa wasl (confusion in conjunction use/non-use) \\
                                        & \textbf{SO}        & Other semantic errors                               \\ \midrule
\multirow{4}{*}{\textbf{Punctuation}}   & \textbf{PC}        & Punctuation confusion                               \\
                                        & \textbf{PT}        & Unnecessary punctuation                             \\
                                        & \textbf{PM}        & Missing punctuation                                 \\
                                        & \textbf{PO}        & Other errors in punctuation                         \\
\textbf{Merge}                          & \textbf{MG}        & Words are merged                                    \\\midrule
\textbf{Split}                          & \textbf{SP}        & Words are split        \\ 
\toprule
\end{tabular}}

\caption{ The ALC error type taxonomy extended with merge and split classes} \label{tab:errors_class}
\end{table*}

\section{ARETA Results}\label{app: areta}
\begin{table*}[!htp]\centering
\scriptsize
\resizebox{\textwidth}{!}{
\begin{tabular}{cccccccc}\toprule
\textbf{CLASS} & \textbf{GECToR\_ARBERT} & \textbf{five-shot\_2014\_expertprompt} & \textbf{five-shot\_2014-chatgpt4} & \textbf{AraT5 (11M)} & \textbf{COUNT} \\\midrule
OH & 73.73 & 89.80 & 92.91 & 87.34 & 4902 \\
OT & 76.59 & 94.12 & 95.58 & 90.84 & 708 \\
OA & 78.63 & 84.66 & 88.93 & 87.35 & 275 \\
OW & 38.57 & 80.79 & 86.96 & 83.70 & 107 \\
ON & 0.00 & 0.00 & 0.00 & 0.00 & 0 \\
OG & 48.00 & 55.74 & 63.64 & 90.32 & 34 \\
OC & 21.43 & 28.57 & 53.66 & 87.18 & 22 \\
OR & 38.24 & 53.02 & 65.96 & 77.10 & 528 \\
OD & 33.76 & 51.89 & 59.60 & 73.07 & 321 \\
OM & 41.80 & 44.53 & 57.35 & 86.44 & 393 \\
OO & 0.00 & 0.00 & 0.00 & 0.00 & 0 \\
MI & 11.02 & 13.25 & 20.53 & 75.00 & 83 \\
MT & 0.00 & 7.84 & 11.43 & 62.50 & 7 \\
XC & 32.95 & 46.10 & 50.78 & 88.35 & 526 \\
XF & 6.06 & 17.98 & 23.81 & 76.92 & 29 \\
XG & 37.10 & 19.57 & 31.35 & 89.47 & 79 \\
XN & 25.19 & 25.79 & 31.25 & 88.12 & 108 \\
XT & 3.95 & 3.78 & 5.48 & 2.48 & 66 \\
XM & 2.04 & 4.14 & 6.38 & 1.07 & 26 \\
XO & 0.00 & 0.00 & 0.00 & 0.00 & 0 \\
SW & 50.51 & 21.25 & 33.38 & 8.29 & 219 \\
SF & 0.00 & 6.67 & 3.45 & 57.14 & 3 \\
PC & 60.89 & 56.25 & 47.59 & 74.98 & 713 \\
PT & 29.62 & 29.58 & 21.40 & 57.42 & 480 \\
PM & 55.24 & 54.21 & 52.09 & 67.08 & 5599 \\
MG & 25.05 & 75.96 & 79.70 & 64.80 & 434 \\
SP & 42.27 & 90.93 & 91.61 & 86.70 & 805 \\\midrule
\textbf{micro avg} & 55.67 & 60.05 & 64.51 & 57.28 & 16467 \\
\textbf{macro avg} & 30.84 & 39.13 & 43.51 & 61.62 & 16467 \\
\textbf{weighted avg} & 56.98 & 66.96 & 68.24 & 76.35 & 16467 \\
\bottomrule
\end{tabular}}
\caption{ Analysis of Error Type performances on the QALB-2014 Test set.}\label{tab:model_performance_comparison}
\end{table*}

\section{Dev Results}\label{app: 2014 dev Results}
\begin{table*}[!htp]\centering
\scriptsize
\resizebox{\textwidth}{!}{
\begin{tabular}{lrrrrrrrrrrrrrrrrrr}\toprule
\multirow{2}{*}{\textbf{Settings}} &\multirow{2}{*}{\textbf{Models}} &\multicolumn{4}{c}{\textbf{Exact Match}} &\multicolumn{4}{c}{\textbf{No Alif / Ya Errors}} &\multicolumn{4}{c}{\textbf{No Punctuation}} &\multicolumn{4}{c}{\textbf{No Puncation and Alif / Ya Errors}} \\\cmidrule{3-18}
& &\textbf{P} &\textbf{R} &\textbf{F\textsubscript{1.0}} &\textbf{F\textsubscript{0.5}} &\textbf{P} &\textbf{R} &\textbf{F\textsubscript{1.0}} &\textbf{F\textsubscript{0.5}} &\textbf{P} &\textbf{R} &\textbf{F\textsubscript{1.0}} &\textbf{F\textsubscript{0.5}} &\textbf{P} &\textbf{R} &\textbf{F\textsubscript{1.0}} &\textbf{F\textsubscript{0.5}} \\\midrule
\multirow{4}{*}{\textbf{Encoder-Only }} &ARBERTv2 &73.30 &47.85 &57.90 &66.25 &65.60 &44.20 &52.81 &59.81 &72.38 &48.75 &58.26 &65.98 &57.40 &33.90 &42.63 &50.41 \\
&ARBERTv2 3 Stage &74.65 &46.70 &57.46 &66.67 &65.00 &41.20 &50.43 &58.27 &75.50 &44.50 &56.00 &66.27 &55.70 &27.50 &36.82 &46.22 \\
&MARBERTv2 &72.95 &47.65 &57.65 &65.95 &64.60 &43.20 &51.78 &58.78 &73.72 &44.16 &55.23 &65.02 &56.80 &34.20 &42.69 &50.17 \\
&MARBERTv2 3 Stage &74.55 &45.75 &56.70 &66.21 &65.10 &41.30 &50.54 &58.37 &75.41 &45.52 &56.77 &66.66 &56.00 &29.20 &38.38 &47.31 \\\cmidrule{2-18}
\multirow{5}{*}{\textbf{Decoder-Only}} &LLama 7B Original &58.20 &32.50 &41.71 &50.25 &35.50 &16.70 &22.71 &28.98 &19.60 &54.30 &28.80 &22.47 &65.10 &32.00 &42.91 &53.94 \\
&Alpaca 7B &42.20 &31.20 &35.88 &39.42 &42.20 &33.40 &37.29 &40.09 &82.20 &62.20 &70.81 &77.23 &62.20 &39.50 &48.32 &55.79 \\
&Vicuna 13B &63.90 &51.00 &56.73 &60.82 &51.40 &39.30 &44.54 &48.42 &83.90 &73.90 &78.58 &81.69 &68.50 &49.00 &57.13 &63.45 \\
&bactrian-x-bloom-7b1-lora &60.80 &43.80 &50.92 &56.42 &53.70 &41.00 &46.50 &50.57 &79.40 &63.00 &70.26 &75.47 &62.00 &51.00 &55.96 &59.44 \\
&bactrian-x-llama-7b-lora &58.60 &41.40 &48.52 &54.10 &51.00 &38.10 &43.62 &47.77 &77.00 &59.20 &66.94 &72.63 &58.60 &48.10 &52.83 &56.15 \\\cmidrule{2-18}
\multirow{20}{*}{\parbox[c]{1in}{\centering\textbf{Encoder Decoder Models}}} &mT0 &69.35 &54.29 &60.90 &65.70 &57.45 &42.50 &48.86 &53.67 &82.35 &75.34 &78.69 &80.85 &70.20 &50.30 &58.61 &65.05 \\
&mT5 &69.00 &53.20 &60.08 &65.13 &56.70 &39.50 &46.56 &52.16 &81.00 &70.00 &75.10 &78.53 &68.00 &48.00 &56.28 &62.77 \\
&AraBART &72.00 &61.50 &66.34 &69.62 &60.00 &49.70 &54.37 &57.61 &85.00 &78.50 &81.62 &83.62 &74.00 &60.50 &66.57 &70.84 \\
&AraT5 &74.50 &64.50 &69.14 &72.26 &63.50 &52.70 &57.60 &61.00 &88.00 &84.50 &86.21 &87.28 &81.50 &69.50 &75.02 &78.78 \\\cmidrule{2-18}
\bottomrule
\end{tabular}}
\caption{Dev Set results on the QALB-2014 benchmark dataset.}\label{tab: dev}
\end{table*}

\end{document}